\documentclass[10pt,twocolumn,letterpaper]{article}
\usepackage[pagenumbers]{wacv} 

\definecolor{wacvblue}{rgb}{0.21,0.49,0.74}
\usepackage[pagebackref,breaklinks,colorlinks,allcolors=wacvblue]{hyperref}

\def\wacvPaperID{2260} 
\def\confName{WACV}
\def\confYear{2026}

\title{Latent Beam Diffusion Models for Generating Visual Sequences}

\author{
 \textbf{Guilherme Fernandes\textsuperscript{1}},
 \textbf{Vasco Ramos\textsuperscript{1}},
 \textbf{Regev Cohen\textsuperscript{2}},
 \textbf{Idan Szpektor\textsuperscript{2}},
 \textbf{Joao Magalhaes \textsuperscript{1}}
\\
 \textsuperscript{1}NOVA LINCS, NOVA School of Science and Technology, Portugal
 \\
 \textsuperscript{2}Google Research
\\
\email{jmag@fct.unl.pt},
\email{szpektor@google.com}
}

\usepackage[utf8]{inputenc} 
\usepackage[T1]{fontenc}    
\usepackage{hyperref}       
\usepackage{url}            
\usepackage{booktabs}       
\usepackage{amsfonts}       
\usepackage{nicefrac}       
\usepackage{microtype}      
\usepackage{xcolor}         
\usepackage{graphicx}         
\usepackage{amsmath}
\usepackage{wrapfig}
\usepackage{enumitem}
\usepackage{algorithm}
\usepackage{algorithmic}
\usepackage{tcolorbox}

\newcommand{\model}{BeamDiffusion\xspace}
\newcommand*{\email}[1]{\texttt{#1}}

\newcommand*{\inlineequation}[2][]{
  \begingroup
    \refstepcounter{equation}%
    \ifx\\#1\\%
    \else
      \label{#1}%
    \fi
    \relpenalty=10000 %
    \binoppenalty=10000 %
    \ensuremath{%
      #2%
    }%
    ~\@eqnnum
  \endgroup
}

\DeclareMathOperator*{\argminB}{argmin}   

\newcommand{\argmaxE}{\mathop{\mathrm{argmax}}}          
\usepackage{caption}
\usepackage{multirow, makecell}
\usepackage{rotating}
\usepackage{float}
\usepackage{caption}
\usepackage{subcaption}
\usepackage{mathabx}
\usepackage{xspace}

\newcommand\tab[1][0.5cm]{\hspace*{#1}}

\begin{document}

\maketitle

\begin{abstract}
While diffusion models excel at generating high-quality images from text prompts, they struggle with visual consistency when generating image sequences. Existing methods generate each image independently, leading to disjointed narratives — a challenge further exacerbated in non-linear storytelling, where scenes must connect beyond adjacent images.
We introduce a novel beam search strategy for latent space exploration, enabling conditional generation of full image sequences with beam search decoding. In contrast to earlier methods that rely on fixed latent priors, our method dynamically samples past latents to search for an optimal sequence of latent representations, ensuring coherent visual transitions. As the latent denoising space is explored, the beam search graph is pruned with a cross-attention mechanism that efficiently scores search paths, prioritizing alignment with both textual prompts and visual context.
Human and automatic evaluations confirm that \model outperforms other baseline methods, producing full sequences with superior coherence, visual continuity, and textual alignment. \footnote{Data and Code: \href{https://huggingface.co/Gui28F/BeamDiffusion}{huggingface.co/Gui28F/BeamDiffusion}}

\end{abstract}

\section{Introduction}
Image diffusion models~\cite{sohl2015deep, ho2020denoising, song2020denoising} have made significant advancements in generating high-quality images. These models, especially when combined with text prompts, have shown great potential in producing detailed and contextually accurate visuals~\cite{rombach2022high, podell2023sdxlimprovinglatentdiffusion,saharia2022photorealistictexttoimagediffusionmodels}. Despite the impressive results at generating individual images based on specific text prompts, diffusion models face challenges when it comes to creating coherent sequences of images. In most cases, each image is generated independently based on its own prompt, which does not naturally ensure visual continuity across a \textit{sequence of correlated prompts}. This problem is further compounded when the narrative is non-linear, where a scene may be connected not only to the immediate prior scene but also to earlier scenes in the sequence. 
\begin{figure}
    \centering
    \includegraphics[clip, trim=70 0 70 0,width=\linewidth]{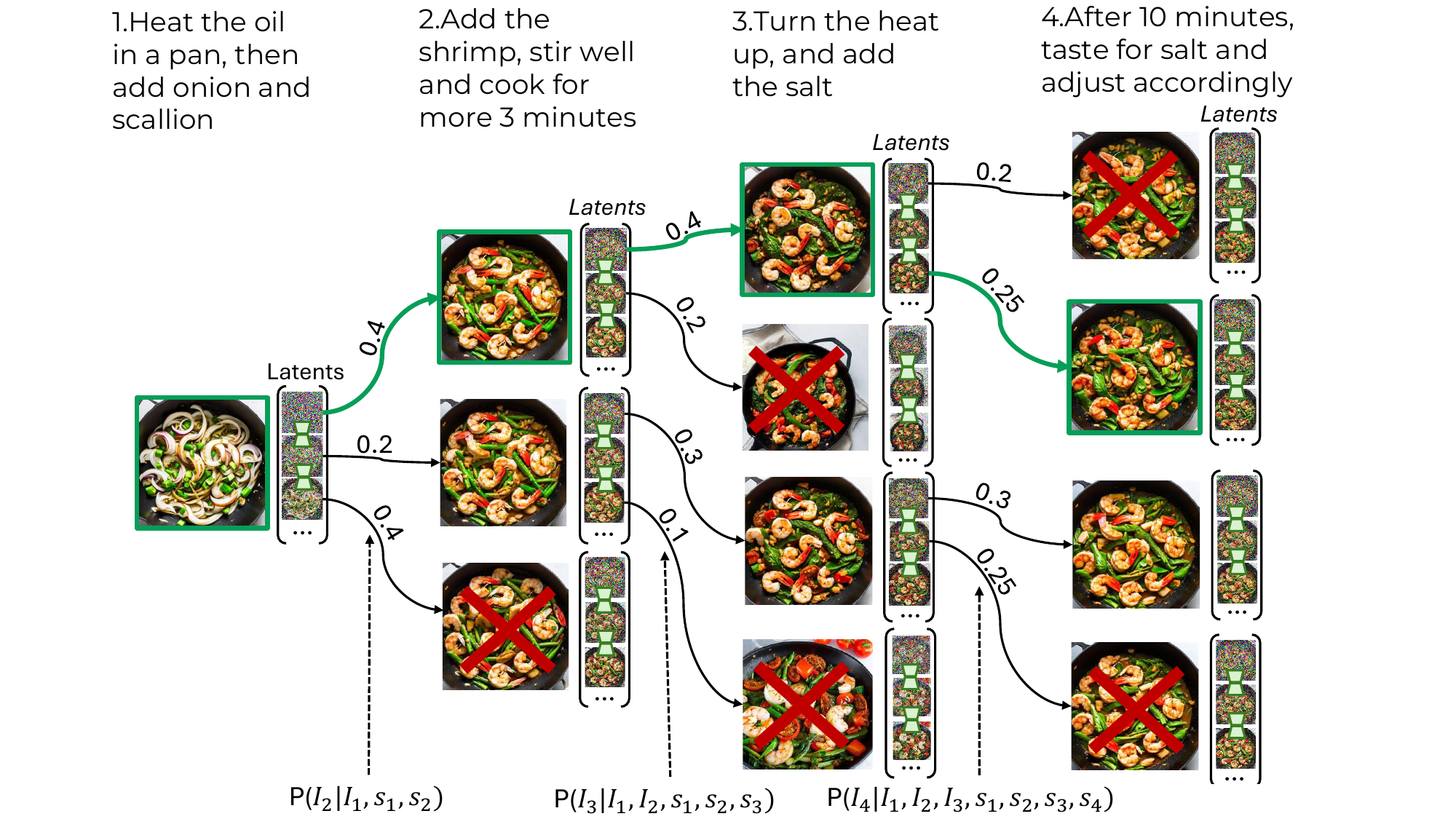}
    \caption{\model leverages search in the latent denoising space to generate a coherent sequence of images from stepwise text instructions. The model conditions the generation of each new image \(I_j\) on latents from previous steps \(s_j\), forming a chain that carries forward information from past generations. At each step, multiple candidate latents are explored, the worst candidates are discarded, and the best ones are kept for the next step.}
    \label{fig:conceptual_idea}
\end{figure}

We propose to research beam search as a scientifically well-grounded solution to this problem. Beam search is a core text generation technique for tasks like machine translation and standard inference~\cite{kasai2024call, wu2016googlesneuralmachinetranslation, KoehnK17, RushCW15}, allowing the model to explore multiple possible paths during the generation process, evaluating different trajectories and refining its predictions. 
However, despite its success in other domains, it has been barely explored in this context, leaving its potential untapped. 
In full image sequence generation, beam search could help navigate the complex latent space, refining coherence by adjusting to multiple text prompts or adding semantic cues, as illustrated in Figure~\ref{fig:conceptual_idea}.

\begin{figure*}[t]
    \centering
    \begin{subfigure}[b]{0.97\textwidth}
        \includegraphics[clip=true, trim= 0 49 0 160, width=\textwidth]{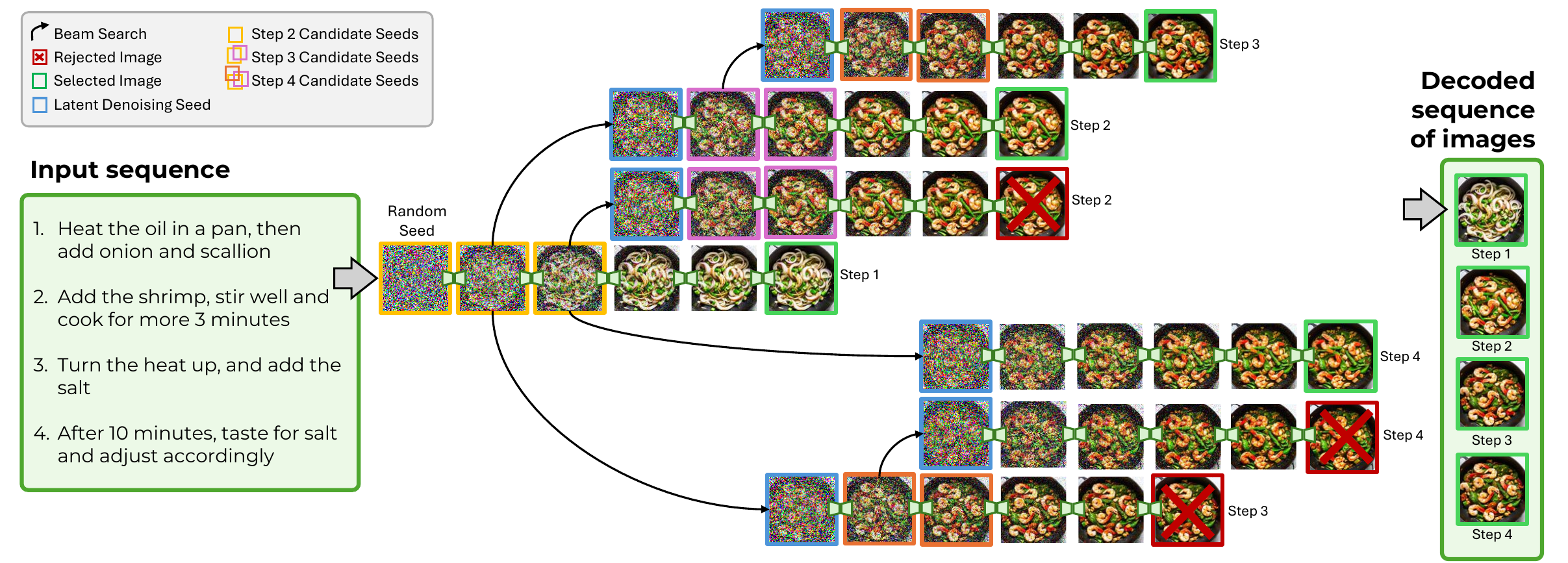}
        \caption{The beam denoising search tree.}
        \label{fig:arch}
    \end{subfigure}
    \vspace{0.4cm}
    \\
    \begin{subfigure}[b]{0.70\textwidth}
        \includegraphics[clip=true,trim= 270 220 200 270, width=\textwidth]{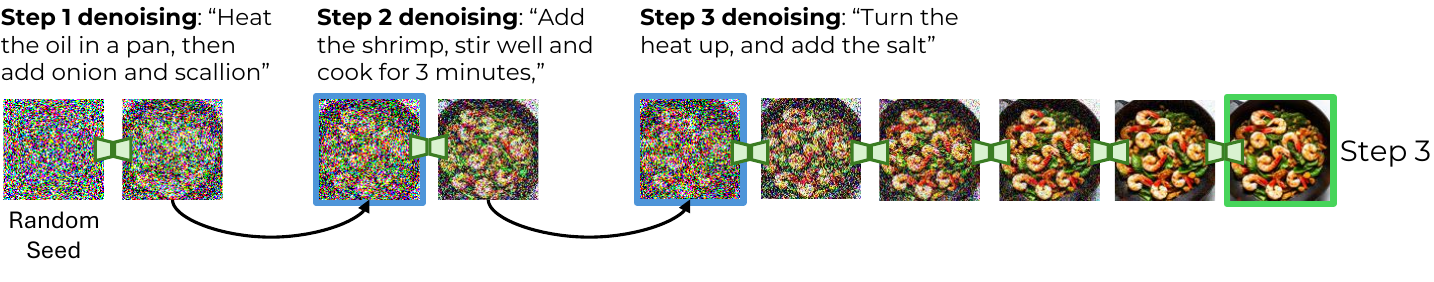}
        \caption{Example of one complete decoded denoising process using contextual captions.}
        \label{fig:img_gen}
    \end{subfigure}
    \caption{The \model model works in the denoising latent space by denoising the initial seed to latents and them decoding different beam paths. This shows how our method evolves in the latent space and how it can explore the latent space.}
    \label{fig:comp_arch}
\end{figure*}

In this work, we revisit the latent diffusion framework with a focus on structured sequence generation. We aim to dynamizally sample the latent space of past diffusion processes and propose the BeamDiffusion method for improving multi-image visual consistency by leveraging shared latent representations in a principled way.
BeamDiffusion explores the latent denoising trajectories with a beam search strategy for ensuring a conditional dependency between diffusion processes. 
As illustrated in Figure~\ref{fig:arch}, our approach leverages a sampling strategy in the latent space to identify the best sequence of images, minimizing the risk of falling into suboptimal or incoherent paths, while maintaining visual continuity across images. 
As the beam diffusion algorithm traverses the latent space, the number of hypothesis grows and the search paths with the lowest probability are pruned to keep a fixed number of candidate paths. 
To this end, we incorporate a cross-attention mechanism~\cite{lin2022cat} that evaluates and scores candidate image paths based on their alignment,
providing a reliable quality measure for each path, allowing us to retain the best ones while pruning the least likely.

To test the generalization and robustness of \model, we evaluated it in two diffusion architectures (LDM~\cite{rombach2022high} and DiT~\cite{blackforestlabs_flux}) and two domains (recipes and visual storytelling domains). 
Results obtained with human annotations, automatic evaluations and LLM-as-a-judge, all showed that the proposed approach outperforms baseline methods for generated sequence quality, showcasing its adaptability to diverse contexts.

\section{Related Work}
Decoding methods are crucial for enhancing the quality of large language models (LLMs). Techniques such as beam search, top-k and nucleus sampling are widely studied for their effectiveness in balancing computational cost with generating coherent, contextually appropriate outputs \cite{shi2024thoroughexaminationdecodingmethods,holtzman2020curiouscaseneuraltext,furniturewala2024impactdecodingmethodshuman}.
Beam search has also proven effective in tasks like machine translation~\cite{machineTranslation}, image captioning~\cite{Vijayakumar_Cogswell_Selvaraju_Sun_Lee_Crandall_Batra_2018,reflective_decoding}, and visual storytelling~\cite{hsu2018usingintersentencediversebeam}. 

Ensuring coherence remains a major challenge in multimodal sequence generation \cite{geyer2023tokenflow,liu2023syncdreamer,chen2024id,khachatryan2023text2video}. Some approaches, like AR-LDM~\cite{AR-LDM}, encode the context of caption-image pairs into multimodal representations to maintain consistency across outputs. Likewise, \citet{make-a-story} incorporates the history of U-net~\cite{ronneberger2015unetconvolutionalnetworksbiomedical, salimans2017pixelcnn++} latent vectors to preserve narrative coherence. 
Other approaches, such as Gill~\cite{gill}, combine LLMs with image encoder-decoder systems. StoryDiffusion~\cite{storydiffusion}, changes the U-Net's cross-attention mechanism enabling training-free consistency by sampling tokens from previous images to maintain coherence in newly generated frames. StoryGen~\cite{StoryGen} denoises the previous image and the resulting features serve as conditioning for the subsequent generation of images. GenHowTo~\cite{GenHowTo} uses a control net during inference to project the previous image directly into the latent space to guide the diffusion process for the next image step. 
StackedDiffusion~\cite{StackedDiffusion} processes stacked image sequences as a single latent representation allowing the attention mechanism to consider all steps together, enhancing coherence across steps. In \cite{bordalo24}, authors proposed to use latent information from previous steps to guide coherent generation of subsequent images relying on a greedy heuristic-based selection. CoSeD~\cite{CoSeD} improved on this idea and introduced a contrastive learning approach~\cite{chen2020simple} to rank images generated from different latent seeds. However, both approaches focused on local refinements rather than optimizing throughout the sequence. Our beam search approach extends these local selection criteria to perform global optimization.

While effective for certain tasks, these models suffer from alignment issues, with retrieved or generated images failing to match the narrative context, and they too often require complex training. Our method addresses these previous constraints by generating images directly from the guiding text, maintaining consistency via a beam search for optimal seeds, similar to methods such as~\cite{inf_time_steering_diff} that view image generation as a search process.

\section{Generating Non-Linear Visual Sequences}
\label{sec:3}

Generating a full sequence of visually coherent images from textual scene descriptions is a challenging research problem. Given a sequence of steps \( S = \{s_1, s_2, \dots, s_L\} \), where each sequence step \( s_j \) is a textual description, our goal is to generate a corresponding sequence of images \( \{I_1, I_2, \dots, I_L\} \) that maintain consistency across all steps of the sequence.  
Diffusion Models generate high-quality images in a compressed latent space, but typically treat each image independently from a randomly initialized latent \( z_T \).

Within this framework, Latent Diffusion Models (LDMs)~\cite{rombach2022high} and Diffusion Transformers (DiTs)~\cite{DiT} share a common structure: both encode an input image \( I \) into a latent representation \( z \) using an encoder \( \mathcal{E} \) and reconstruct it through a decoder \( \mathcal{D} \)~\cite{rombach2022high}. The reverse diffusion process is performed by a denoising backbone \( \epsilon_\theta \), which can be a U-Net~\cite{ronneberger2015unetconvolutionalnetworksbiomedical} (for LDMs) or a transformer~\cite{DiT} (for DiT), and incorporates cross-attention to condition the denoising iterations on an external input, e.g., the output embeddings of a text encoder \( \tau_\theta(s) \). Both models are trained to iteratively estimate a denoised latent with the loss:
\begin{equation}
L_{\text{diff}} = \mathbb{E}_{\mathcal{E}(I), s, \epsilon \sim \mathcal{N}(0,1), t} \left[ \| \epsilon - \epsilon_\theta(z_t, t, \tau_\theta(s)) \|_2^2 \right]
\end{equation}

This formulation highlights that each denoising process is \emph{agnostic to other generations}, making it challenging to maintain coherence across a sequence of related prompts. This is particularly important in non-linear narratives~\cite{smilevski2018stories}, where preserving character identity, lighting, and environmental details is crucial. To address these challenges, prior work~\cite{make-a-story, StoryGen, GenHowTo, bordalo24, CoSeD} has proposed \emph{reusing latent denoising iterations from earlier steps}, raising the key question: \textit{how can we determine the optimal set of latent states that maximizes coherence across different generations?} 

In the following subsections, we introduce \model (Sec.~\ref{sec:beam_diffusion}), detail latent space exploration with diffusion beams (Sec.~\ref{sec:diffusion_beams_hypothesis}), and outline our beam pruning strategy (Sec.~\ref{sec:pruning_diffusion_beam_paths}).

\subsection{Beam Diffusion Model}
\label{sec:beam_diffusion}
To address the limitations of standard diffusion models in generating coherent image sequences, we introduce the \model model. The proposed approach aims to enhance both the consistency and quality of image sequences by adopting a strategy inspired by beam search, but with a twist, it operates on latent denoising space. 
\model incorporates latent representations from previously generated images, Figure~\ref{fig:arch}. These latent representations guide the generation of new candidates, ensuring that each new image aligns with the context of the evolving sequence. This approach allows visual elements from previous images to be carried over or referenced, maintaining contextual consistency and continuity, while also adapting to the next scene description.

In this article, the term \textit{step} will refer to an element in the input sequence; and the term \textit{iteration} will refer to one denoising iteration of the diffusion model.

\subsection{Latent Space Exploration with Diffusion Beams}
\label{sec:diffusion_beams_hypothesis}
To generate coherent and diverse image sequences, we explore the latent space of diffusion models using a beam search-inspired strategy. For each sequence step $j$ we generate a set of candidate images \( \mathcal{I}_j =\{I_j^{(1)},I_j^{(2)}, \dots \} \), where each \( I_j^{(k)} \) is conditioned on a different beam path.

\paragraph{Sampling the Denoising Latent Space.} 
  
In our approach, the denoising process for each sequence step is conditioned by representations from prior steps, helping the model maintain coherence across the sequence.
In the first step of the sequence, \( s_1 \), we generate multiple candidate images by running the diffusion process with  $r$ random noise seeds (Alg.~\ref{alg:beamdiff}, line 2). This strategy enables broader exploration of the latent space and reduces Bayes risk~\cite{bertsch-etal-2023-mbr}, as detailed in Appendix~\ref{appendix:minimizing_decoding_risks}. 

For every generation process corresponding to a sequence step $s_i$, we store the first $N$ latent space vectors $z_i = \{ z_{i,t} \}_{t \in \subseteq \{1, \ldots, N\}}$ produced during the denoising iterations. 
On each step prompt $s_{j>1}$ we need to generate diverse samples that (a) explore different denoising trajectories in the latent space, and (b) are strongly correlated with previous samples. 
Leveraging the stored latent space vectors, we sample the latent subspace of $n$ previous denoising processes. 
Hence, for steps \(j > 1\), to enable contextual consistency, we first accumulate representations from the previous \( m \) steps:
\begin{equation}
    \label{eqn:latents_set}
    \textstyle 
    \mathcal{Z}_j = \bigcup_{i \in \{j-m, \ldots, j-1\} } \{z_i\}.
\end{equation}

To generate new candidate images \( \mathcal{I}_j = \{ I_j^{(1)}, \ldots, I_j^{(k)}, \ldots, I_j^{(N \cdot m)} \} \) at step \( j \), we condition the diffusion model on the latent trajectories in \( \mathcal{Z}_j \). Specifically, for each latent trajectory \( z \in \mathcal{Z}_j \), the model performs a new diffusion process to produce a corresponding image candidate \( I_j^{(k)} \), Alg.~\ref{alg:beamdiff}, line 5 and 7. This step establishes a dependency between generations, where latents from earlier steps guide the synthesis of visually coherent transitions over time, as illustrated in Figure~\ref{fig:comp_arch}.

\begin{algorithm}[t]
\small
\caption{BeamDiffusion Algorithm}
\label{alg:beamdiff}
\begin{algorithmic}[1]
\STATE \textbf{Input parameters:} \\$\{s_1,\ldots, s_L\}$: sequence of steps; $r$: number of random seeds; $m$: steps back; $w$: beam width.
\STATE $\hat{B}_{1} \leftarrow \{\text{SD}(s_1, rnd)\}^r$ \hfill $\triangleright$ Beam initialization with $r$ different images.
\FOR{$j = 2$ \TO $L$}
    \FORALL{beam $b \in \hat{B}_{j-1}$}
        \STATE $\mathcal{Z}_j \leftarrow$ $\{\hat{B}_{j-n}, \ldots, \hat{B}_{j-1}\}$ \hfill $\triangleright$ Read denoising latents from previous steps.
        \FORALL{latent $z \in \mathcal{Z}_j$}
            \STATE $I_j \leftarrow \text{SD}(s_j, z)$ \hfill $\triangleright$ Generate candidate a image.
            \STATE $\hat{B}_{j-1} \leftarrow \hat{B}_{j-1} \bigcup {I_j}$ \hfill $\triangleright$ Extend beam with candidate image.
        \ENDFOR
    \ENDFOR
    \STATE $\text{score}(I_j) \leftarrow \varphi(I_j \in \hat{B}_j)$  \hfill $\triangleright$ Score the new image of the beam.
    \STATE $B_j \leftarrow \arg\max_{|B_j|=w} \text{score}(\hat{B}_j)$ \hfill $\triangleright$ Prune beams and keep top $w$ beams. 
\ENDFOR
\RETURN best beam $B^* \leftarrow \arg\max_{B_L} \text{score}(B_L)$
\end{algorithmic}
\end{algorithm}

\paragraph{Diffusion Beams.} 
Inspired by beam search~\cite{wiseman2016sequence}, which maintains and extends the most promising candidates at each sequence step, we introduce \emph{diffusion beams} to structure sequential image generation. At each sequence step \( j \), the set of candidate images \( \mathcal{I}_j \) is used to compute the set of candidate diffusion beams \( \hat{B}_j \):
for each step \( j \) we consider all the beams existing in step $j-1$ and extend each one of such $b$ beams \( \hat{B}_{j-1}^{(b)} \) by pairing it with a new candidate image \( I_j^{(k)} \in \mathcal{I}_j \). The resulting beams are defined as 
\begin{equation}
    \textstyle 
\hat{B}_j = \bigcup_{(b, k)} \left\{ \hat{B}_{j-1}^{(b)} \oplus I_j^{(b \cdot k)} \mid I_j^{(b \cdot k)} \in \mathcal{I}_j \right\},
\end{equation}
where \( \oplus \) denotes the concatenation of a prior beam with a new candidate image, and \( (b \cdot k) \) indexes the pairing of beam \( b \) and candidate image \( I_j^{(k)} \). This process builds a tree of candidate sequences, Alg.~\ref{alg:beamdiff}, line 8, enabling the model to explore diverse yet coherent visual narratives over time.

\subsection{Pruning Diffusion Beam Paths}
\label{sec:pruning_diffusion_beam_paths}

Tracking multiple \emph{diffusion beams} enables \model to maintain visual coherence while exploring diverse continuations across a sequence. However, as the sequence length increases, the number of candidate beams grows rapidly, making it computationally infeasible to retain them all. To address this, we apply a pruning strategy based on a learned contrastive scoring function, which allows us to retain only the most promising beams while preserving diversity.

At each sequence step \( j \), we construct a set of candidate beams \( \hat{B}_j \). Each candidate beam \( \hat{B}_j^{(b)} \) is formed by extending a beam from the previous step \( \hat{B}_{j-1}^{(b)} \) with a newly generated image \( I_j^{(k)} \). A complete candidate beam at step \( j \) is a sequence of images \( \hat{B}_j^{(b)} = [I_1^{(b)}, I_2^{(b)}, \ldots, I_j^{(b)}] \), one per step.
To evaluate the overall quality of a beam, we use a contrastive classifier~\cite{CoSeD}, denoted by \( \varphi \). Specifically, for each image \( I_i^{(b)} \) in the beam, the classifier \(\varphi\) computes a compatibility score based on the image's prompt \( s_i \) and the previous images in the beam \( B_{i-1}^{(b)} \), which includes earlier images and prompts in the same beam. The total beam score is then the sum of these per-image scores over all sequence steps up to \( j \):
\begin{equation}
\label{eq:beam_score}
\textstyle 
score(\hat{B}_j^{(b)}) = \sum_{i=1}^{j} \log \left( \frac{ \exp \left(\varphi(I_i^{(b)}, s_i, B_{i-1}^{(b)}) \right) }{ \sum_{I_i'\in \mathcal{I}_i} \exp \left( \varphi(I_i', s_i, B_{i-1}^{(b)}) \right) } \right),
\end{equation}
where \( \mathcal{I}_i \) is the set of all candidate images at sequence step \( i \). The contrastive formulation ensures that each image is evaluated not in isolation, but relative to other candidates, capturing how well it fits the evolving sequence context.

Once scored, we prune the beam candidates by selecting the top \( w \) scoring beams:
\begin{equation}
\label{eqn:beam_width}
\textstyle 
B_j = \argmaxE_{B_j \subseteq \hat{B}_j, |B_j| = w}\left( \sum_{i=1}^w score(\hat{B}_j^{(i)})\right),
\end{equation}
where \( B_j \) denotes the final set of retained beams for sequence step \( j \).

To balance exploration and efficiency, pruning is disabled for the initial steps, allowing all candidates to be retained initially. After a few step, we apply pruning to keep only the top \( w \) beams. This strategy ensures wide exploration early on and focused, high-quality continuation in later sequence steps.

\paragraph{Training.}
The contrastive classifier $\varphi(I_j | \{s_1,\ldots,s_j\}, \{I_1, \ldots, I_{j-1}\} )$ is trained as a \textit{next image prediction} task to estimate how well aligned an image is with the previously sequence of steps and images. For a given sequence \( t \) and sequence step \( j \), the ground-truth label \( l_{t,j} \) indicates whether image \( I_j \) belongs to the sequence given its prompt \( s_j \) and context \( B_{j-1}^{(b)} \). Negative examples are hallucinated from other sequences. The model minimizes the cross-entropy loss
encouraging the model to promote candidates that best match the sequence's semantic and visual dynamics. The classifier is trained on the Recipes~\cite{bordalo24} and VIST~\cite{huang2016visual} datasets. See appendix~\ref{appendix:implementation_details} for implementation details.

\section{Experimental Setup}
\label{sec:4}
In this section, we describe the experimental setup for evaluating \model's performance in generating coherent image sequences. We first outline the different decoding strategies that are considered. We then present the datasets and baseline methods used, and describe the evaluation protocols, which include human judgments, automatic metrics, and LLM-as-a-judge evaluations using Gemini 2.5~\cite{gemini2.5}, all designed to assess semantic and visual consistency across generated sequences.

\paragraph{Decoding strategies.}
Our proposed method, \model, uses a beam search over latent sequences to generate coherent image sequences from textual prompts. To evaluate the effectiveness of this decoding strategy, we also compare it against two alternative approaches applied to the same model: \textbf{Greedy / CoSeD$_{len=1}$}~\cite{CoSeD}, which selects at each step the top-scoring image according to CLIP similarity from the first few latents of the previous step, focusing on immediate best-match selection; and \textbf{Nucleus Sampling}~\cite{topp}, a probabilistic strategy that samples images from the subset whose cumulative CLIP probability exceeds a threshold \(p\), encouraging diversity while still prioritizing high-probability outcomes. 
All strategies are applied to the same sequences of prompts using two underlying diffusion backbones, FLUX.1-dev~\cite{blackforestlabs_flux} and SD2.1~\cite{rombach2022high}, allowing a direct comparison of their ability to maintain sequence-level coherence across different models.

\paragraph{Datasets.}
We used publicly available manual tasks in the Recipes domain~\cite{bordalo24}, along with a small portion of the Visual Storytelling (VIST) dataset~\cite{huang2016visual}, using an equal number of tasks from each. Recipes provide detailed, step-by-step instructions, while the sampled VIST tasks offer additional diversity for training the classifier. This combination allows us to evaluate \model's performance on both familiar and novel task types. Together, these datasets enable a thorough assessment of sequence coherence, semantic alignment, and visual consistency. More details can be found in Appendix~\ref{appendix:datasets}.




\paragraph{Baselines.}
To assess the performance of \model, we compared against several competitive baselines that aim to generate coherent image sequences from text: \textbf{GILL~\cite{gill}}, \textbf{StackedDiffusion~\cite{StackedDiffusion}}, \textbf{StoryDiffusion~\cite{storydiffusion}}, and \textbf{One-Prompt-One-Story (1P1S)~\cite{1P1S}}.

\paragraph{Human Evaluation.}

To find the optimal hyperparameters, the annotators compared pairs of beam search configurations to select the optimal beam width ($w$ in Eq.~\ref{eqn:beam_width}) and steps back ($m$ in Eq.~\ref{eqn:latents_set}). 
The winning configuration is selected iteratively by comparing different configurations and preserving always the better one.
See Appendix~\ref{appendix:beam_config} for the detailed results.

Having chosen the optimal configuration, we compared \model to other methods. In the second annotation experiment, \model was compared with other baseline models. Annotators independently assessed the sequences, selecting the best method according to semantic and visual consistency with the possibility of choosing multiple options or none. For further information, see Appendix~\ref{appendix:annotations}.

\paragraph{LLM-as-a-judge Evaluation.}
To further validate our findings and annotate a wider range of tasks, we adopt the LLM-as-a-judge paradigm, leveraging Gemini, to evaluate both configurations and baselines. Its selection process closely mirrored that of human annotators. For details, please refer to Appendix~\ref{appendix:annotations}.

\paragraph{Human and LLM-as-a-judge Agreement.}
We conducted an alignment assessment to measure agreement between human annotators and the LLM-as-a-judge. Seven annotators rated 16 Recipes and 10 VIST tasks, which we compared to the model’s evaluations. To quantify the agreement between the two, we computed Fleiss' kappa~\cite{agreement}, obtaining a value of 70\%, which indicates a substantial level of agreement. Building on this strong alignment, we employed the LLM-as-a-judge to evaluate the remaining sequences.

\paragraph{Automatic Metrics.}

Following prior work~\cite{bordalo24, CoSeD,StoryGen}, we evaluate each method using two complementary criteria: (1) \textit{Visual Consistency}, measured via average image similarity using DINO~\cite{dinov2} (DINO-I) and CLIP~\cite{clip} (CLIP-I); and (2) \textit{Semantic Alignment}, computed with CLIP (CLIP-T) by comparing each image to the average of all step texts up to and including that step. To ensure comparability across metrics, we normalize CLIP-I, CLIP-T, and DINO-I by their respective maxima. We further define CLIP* as the product of CLIP-I and CLIP-T, and DINO* as the product of DINO-I and CLIP-T, balancing visual and semantic quality while emphasizing alignment with the stepwise input text.
To better capture real performance, the results for DINO-I, CLIP-I, CLIP-T, CLIP*, and DINO* are reported as \textit{win percentages} rather than mean values. This avoids overestimating methods that perform very well on some sequences but poorly on others, which could otherwise produce artificially high averages despite inconsistent quality.

In addition, we adopt metrics from StackedDiffusion~\cite{StackedDiffusion} to evaluate task-specific aspects of the generated sequences: \textit{Goal Faithfulness}, measuring alignment of the final image with the overall textual goal; \textit{Step Faithfulness}, measuring alignment of each intermediate image with its corresponding step text; and \textit{Cross-Image Consistency}, capturing visual coherence across images in the sequence. These metrics complement the previous ones by explicitly assessing semantic fidelity and sequence-level consistency. See Appendix~\ref{appendix:annotations} for further details.

\section{Results and Discussion}
\label{sec:5}

In this section, we evaluate the effectiveness of \model through three complementary approaches: human evaluations, LLM-as-a-judge, and automatic metrics. We begin by comparing decoding strategies, selecting the method that achieves the highest human preference scores for subsequent experiments. We then highlight key performance aspects of \model, including visual and semantic consistency, and conclude with a qualitative analysis. Additional details on the annotation setup are provided in Appendix~\ref{appendix:annotations}.

\subsection{Comparison of Decoding Methods}
\label{sec:decoding_comparison}
To evaluate the effectiveness of different decoding strategies, we compare our proposed method, \model, against Greedy/CoSeD$_{len=1}$~\citep{CoSeD} and Nucleus Sampling~\citep{topp} using the same sequences of prompts with two types of diffusion backbones.

\begin{table}[t]
\small
\centering
\caption{Comparison of human (H) and LLM-as-a-judge (L) evaluations of decoding strategies across two backbones.}
\label{tab:decoding_results}
\resizebox{\columnwidth}{!}{%
\begin{tabular}{@{}lcccccc@{}}
    \toprule
     & \multicolumn{2}{c}{\textbf{FLUX.1-dev}} 
     && \multicolumn{2}{c}{\textbf{SD2.1}}\\ 
    \cmidrule(l){2-3} \cmidrule(l){5-6} 
    \textbf{Method} 
    & \textbf{H (\%)} & \textbf{L (\%)} 
    && \textbf{H (\%)} & \textbf{L (\%)}\\ 
    \midrule
    \multicolumn{1}{l|}{Greedy/CoSeD$_{len=1}$~\citep{CoSeD}} 
    & 16.35 & 10.00 
    && 11.11 & 10.00 \\
    
    \multicolumn{1}{l|}{Nucleus Sampling~\citep{topp}} 
    & \underline{24.04} & \underline{16.67} 
    && \underline{33.33} & \underline{26.67} \\

    \multicolumn{1}{l|}{\model} 
    & \textbf{59.61} & \textbf{73.33} 
    && \textbf{55.56} & \textbf{63.33} \\
    \bottomrule
\end{tabular}%
}
\end{table}

As shown in Table~\ref{tab:decoding_results}, \model consistently achieves the highest preference scores according to both human annotators (H) and LLM-as-a-judge (L). In particular, \model substantially outperforms Greedy decoding and Nucleus Sampling, indicating that beam-based decoding produces sequences that are not only more coherent but also better aligned with human and automatic judgments. While Nucleus Sampling improves over Greedy decoding by encouraging diversity, its performance remains notably below that of \model in both evaluation settings. The agreement between H and L further strengthens the evidence for the robustness of our decoding strategy. Given these results, we adopt \model as our default decoding method in subsequent experiments (see Appendix~\ref{appendix:qual_decoding} for qualitative and automatic evaluation results and Appendix~\ref{appendix:beam_config} for hyperparameter configurations).

Table~\ref{tab:decoding_results} also compares two diffusion architectures: Stable Diffusion (SD2.1) and Diffusion in Transformers (FLUX-1-dev). BeamDuffusion exhibits consistent performance with both architecture, with Diffusion in Transformers outperforming SD2.1. Hence, in the following experiments we will use the Diffusion in Transformers architecture.

\begin{table}[t]
\small
\caption{Frequency of method selection by human annotators (H) and the LLM-as-a-judge (L).}
\label{tab:combined-results}
\resizebox{\columnwidth}{!}{%
\begin{tabular}{@{}lcccccc@{}}
    \toprule
     & \multicolumn{2}{c}{\textbf{Recipes}} 
     && \multicolumn{2}{c}{\textbf{VIST}}\\ 
    \cmidrule(l){2-3} \cmidrule(l){5-6} 
    \textbf{Method} 
    & \textbf{H (\%)} & \textbf{L (\%)} 
    && \textbf{H (\%)} & \textbf{L (\%)}\\ 
    \midrule
    \multicolumn{1}{l|}{GILL~\citep{gill}} 
    & 0.00 & 0.00 
    && 2.74 & 0.00 \\
    
    \multicolumn{1}{l|}{StackedDiffusion~\citep{StackedDiffusion}} 
    & 0.00 & 3.33 
    && 0.00 & 0.00 \\
    
    \multicolumn{1}{l|}{StoryDiffusion~\citep{storydiffusion}}
    & \underline{22.58} & \underline{20.00}
    && \underline{36.99} & \underline{41.38}\\
    
    \multicolumn{1}{l|}{1P1S~\citep{1P1S}} 
    & 12.90 & 3.33
    && 19.18 & 10.34 \\
    
    \multicolumn{1}{l|}{BeamDiffusion} 
    & \textbf{64.52} & \textbf{73.33}
    && \textbf{41.10} & \textbf{48.28} \\
    \bottomrule
\end{tabular}%
}
\end{table}

\subsection{General Results}
Table~\ref{tab:combined-results} compares the sequence generation methods on both the Recipes and VIST datasets. \model dominates in both domains, leading Recipes with 64.52\% (H) and 73.33\% (L), and VIST with 41.10\% (H) and 48.28\% (L). 
StoryDiffusion consistently ranks second, with 22.58\% (H) and 20.00\% (L) in Recipes, and 36.99\% (H) and 41.38\% (L) in VIST. 
In contrast, 1P1S receives moderate preference (12.90\% H in Recipes, 19.18\% H in VIST), while GILL and StackedDiffusion record the lowest selection rates.  

These results show that \model achieves the strongest performance across settings, 
with especially large margins in Recipes where \model surpasses the next-best method by over 40\% for human preference. 
The consistently low selection rates of GILL and StackedDiffusion highlight that these methods are rarely preferred by either humans or LLM-as-a-judge, in contrast to the stronger performance of \model and StoryDiffusion. 
Overall, the findings highlight that effective search and selection strategies 
are central to producing coherent and engaging narratives, enabling models 
to maintain quality even in complex domains such as VIST.


\begin{table*}[t]
\small
\centering
\caption{Comparison of different methods across embedding-based sequence quality metrics. CLIP-I, DINO-I, CLIP-T, CLIP*, and DINO* are reported as win percentages, while Goal Faithfulness, Step Faithfulness, and Cross-Image Consistency are reported as absolute scores.}
\label{tab:merged_comparison_all_methods}
\resizebox{\textwidth}{!}{%
\begin{tabular}{@{}lccccc|cccc@{}}
\toprule
\textbf{Method} & \textbf{CLIP-I} & \textbf{DINO-I} & \textbf{CLIP-T} & \textbf{CLIP*} & \textbf{DINO*} & 
\makecell{\textbf{Goal Faith.}$\uparrow$ } &
\makecell{\textbf{Step Faith.}$\uparrow$} &
\makecell{\textbf{Cross-Image Cons.}$\uparrow$} \\ 
\midrule
GILL~\citep{gill} & 0.00 & 0.00 & 0.00 & 0.00 & 0.00 & 0.67 & 0.52 & 0.54 \\
StackedDiffusion~\citep{StackedDiffusion} & \underline{25.00} & \underline{25.00} & 5.00 & \underline{25.00} & \underline{25.00} & 0.60 & 0.67 & \underline{0.57} \\
StoryDiffusion~\citep{storydiffusion} & 20.00 & \underline{25.00} & \underline{35.00} & \underline{25.00} & 20.00 & \underline{0.78} & \underline{0.87} & \textbf{0.61} \\
1P1S~\cite{1P1S} & \underline{25.00} & \textbf{30.00} & 10.00 & 10.00 & \underline{25.00} & 0.75 & 0.83 & \underline{0.57} \\
BeamDiffusion & \textbf{30.00} & 20.00 & \textbf{50.00} & \textbf{40.00} & \textbf{30.00} & \textbf{0.82} & \textbf{0.92} & \textbf{0.61} \\
\bottomrule
\end{tabular}%
}
\end{table*}

\subsection{Automatic Metrics}
\label{sec:automatic_metrics}

Table~\ref{tab:merged_comparison_all_methods} compares all methods using metrics that measure sequence quality, including CLIP-I, DINO-I, CLIP-T, CLIP*, DINO*, and Goal Faithfulness~\cite{StackedDiffusion}, Step Faithfulness~\cite{StackedDiffusion}, and Cross-Image Consistency~\cite{StackedDiffusion}.

\model achieves the highest combined scores (CLIP*: 40.00\%, DINO*: 30.00\%), reflecting strong performance in both visual and semantic consistency. This is further confirmed by faithfulness and consistency metrics, where \model leads in Goal Faithfulness (0.82), Step Faithfulness (0.92), and Cross-Image Consistency (0.61).

StoryDiffusion performs well on CLIP-T (35.00\%) and achieves high scores on the alignment metrics Goal Faithfulness: 0.78, Step Faithfulness: 0.87, Cross-Image Consistency: 0.61, showing strong semantic alignment and visual coherence despite lower combined metrics. 1P1S achieves the highest DINO-I (30.00\%) but lower CLIP-T (10.00\%) and moderate alignment scores, indicating strong visual consistency but weaker semantic alignment. StackedDiffusion maintains balanced performance across CLIP-I, DINO-I, and CLIP* (around 25.00\%) with moderate alignment scores, while GILL consistently underperforms across both automatic and alignment metrics.

Overall, the relationships between CLIP-I/DINO-I and Cross-Image Consistency, and between CLIP-T and Goal/Step Faithfulness, show that the combined metrics CLIP* and DINO* reliably reflect both semantic and visual consistency. Together, these results demonstrate that \model is the most effective method for generating coherent and semantically faithful sequences of images.


\subsection{Non-Linear Sequence Denoising}
\label{sec:non_linear_sequence_denoising}
To analyze how \model explores the latent space, we measured the number of times that a sequence step provides latents for the generation of subsequent steps of the sequence. In this experiment we consider sequences of four steps. 

Latents from step 1 were chosen 69.9\% of the times they had a chance of being selected, highlighting the role of step 1 in shaping the sequence. Selection rates decrease for later steps, with Step 3 at 46.8\%, Step 2 at 33.3\%, and the random seed at 15.6\%. Earlier steps, such as Step 1, have more chances of being chosen, leading to a higher selection rate. Thus, the percentages reflect how often each step is selected given the different possibilities for their inclusion, rather than a simple additive total. This pattern demonstrates how beam search prioritizes crucial steps, particularly Step 1, to establish sequence coherence while strategically weighing intermediate sequence steps.
Appendix~\ref{appendix:latent_selection} provides mode details.

\subsection{Effect of Latent History}
\begin{figure}[t]
    \centering
    \includegraphics[width=0.95\columnwidth]{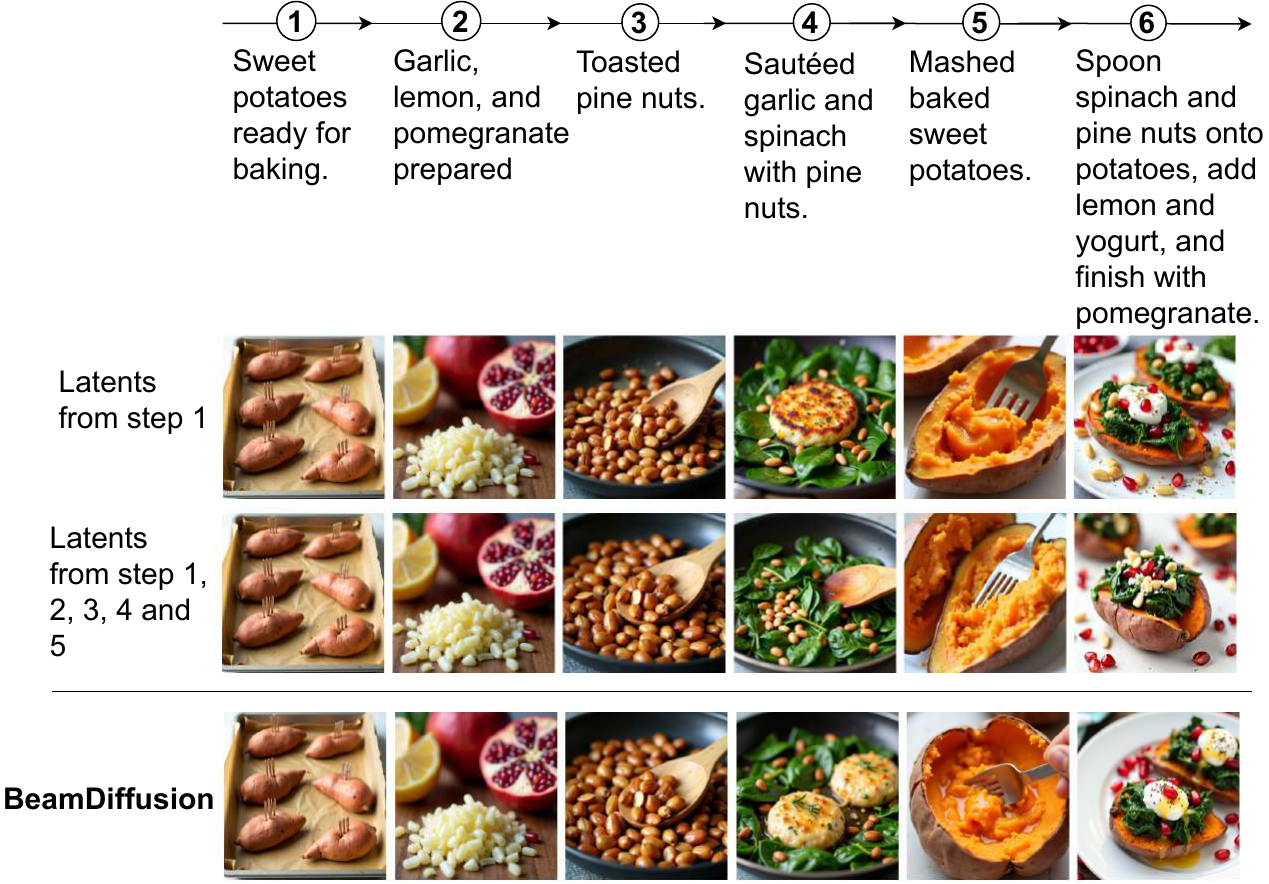} 
    \caption{Visual comparison of sequences generated with cumulative latent access. The first row shows a sequence generated using latents only from step 1, the second row shows a sequence generated using latents from steps 1 through 5, and the bottom row shows the result from \model, which uses latents from only the two preceding steps.}
    \label{fig:cumulative_latents_prev}
\end{figure}
To better understand the role of latent reuse, we compared three settings: using latents only from step 1, using latents from all steps (1–5), and using \model (Figure~\ref{fig:cumulative_latents_prev}). Sequences generated from step 1 alone lacked detail and consistency, while access to all steps improved quality but came with higher computational cost. In contrast, \model, which relies only on the two most recent steps, achieved results comparable to or better than full access while being substantially more efficient. This efficiency arises because each new latent already encodes information from previous steps (Figure~\ref{fig:img_gen}), making full-history conditioning unnecessary. A more detailed analysis of intermediate cases, is provided in Appendix~\ref{appendix:cumulative_latents}.

\begin{figure*}[t]
    \centering
    \begin{subfigure}[t]{0.46\textwidth}
        \centering
        \includegraphics[clip, trim= 30 0 0 0, width=\linewidth]{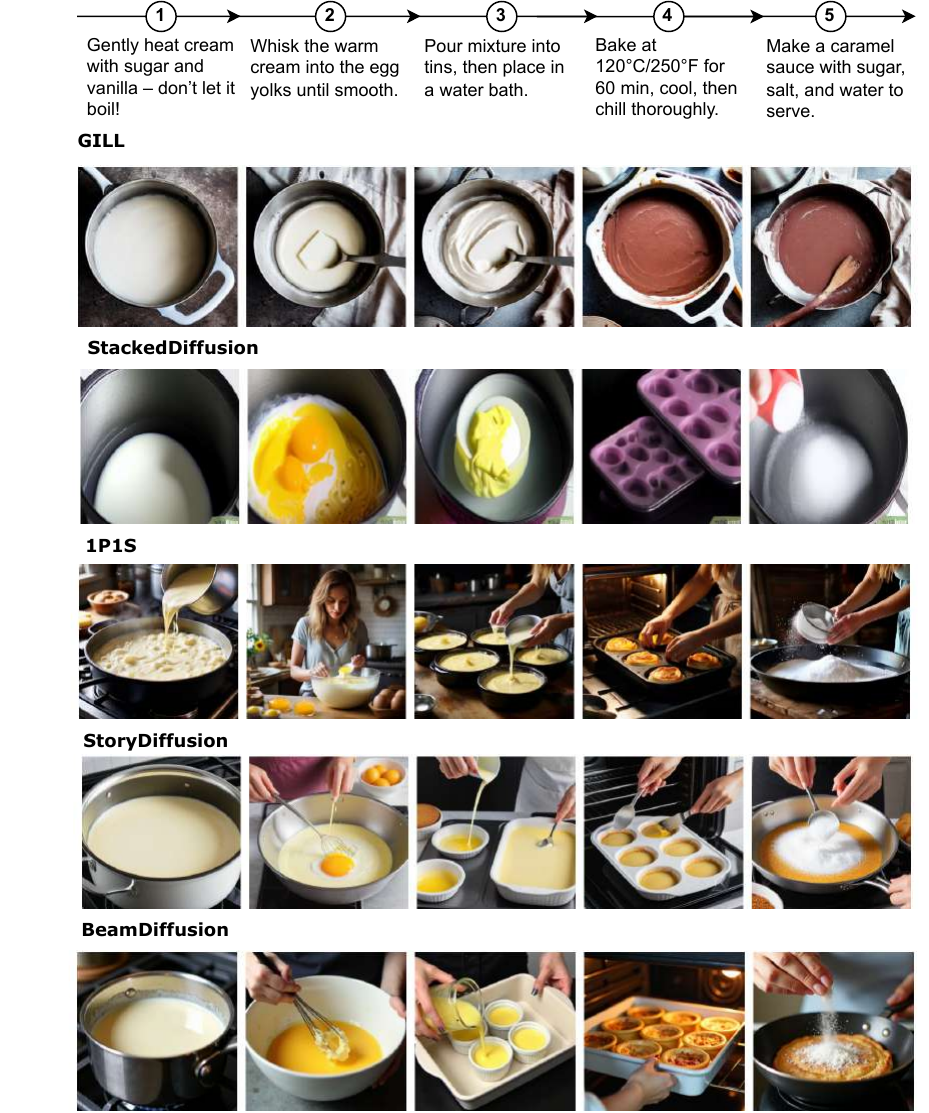}
        \caption{Recipes domain task.}
        \label{fig:recipe_9}
    \end{subfigure}
    \hfill
    \begin{subfigure}[t]{0.46\textwidth}
        \centering
        \includegraphics[clip, trim= 30 0 0 0, width=\linewidth]{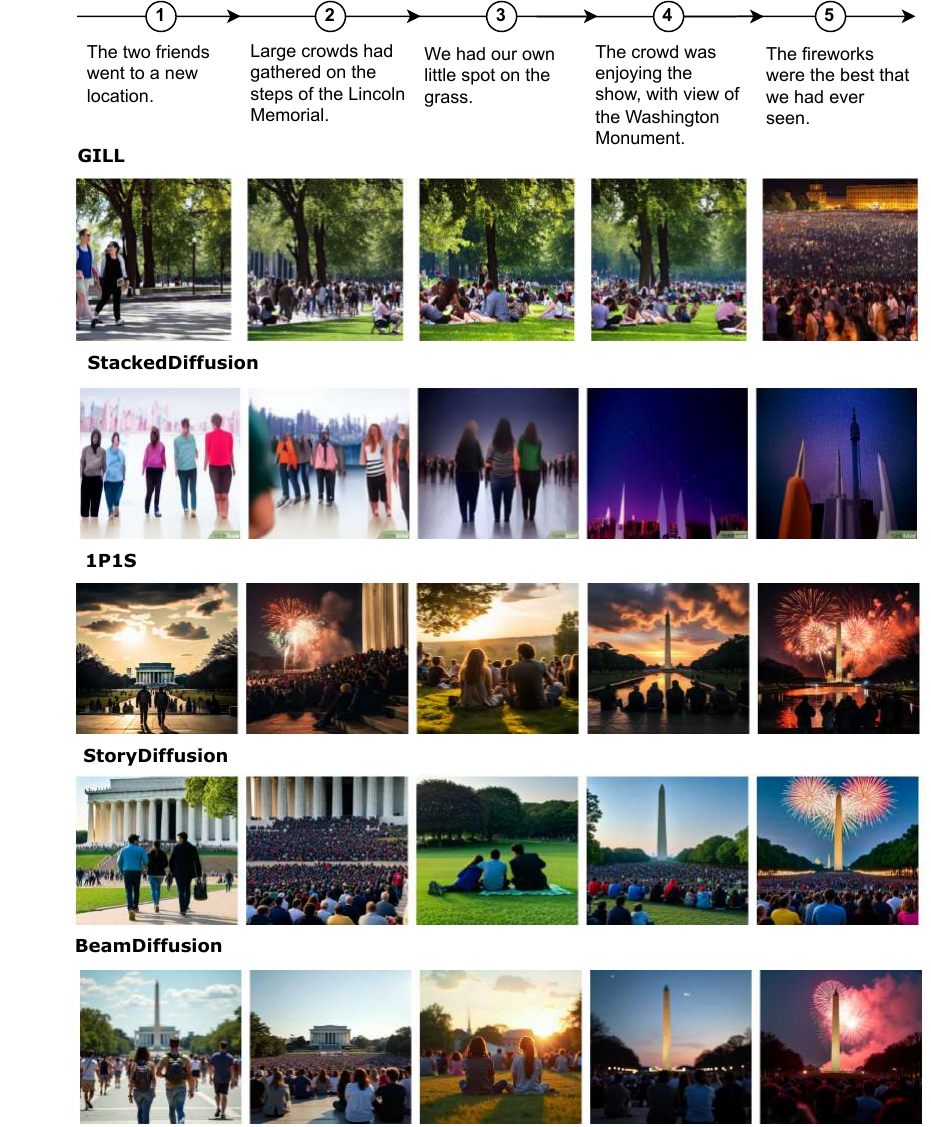}
        \caption{VIST story.}
        \label{fig:vist_0}
    \end{subfigure}
    \caption{Example tasks from two different domains: Recipes (\textit{"Crema Catalana with salted caramel sauce"}) and VIST (\textit{"Fireworks - 4th July in Washington, DC"}).} 
    \vspace{-3mm}
    \label{fig:domain_examples}
\end{figure*}

\subsection{Qualitative Analysis}
To highlight the impact of our approach on semantic and visual consistency, we present two examples, one from the Recipes domain (Figure~\ref{fig:recipe_9}) and one from VIST (Figure~\ref{fig:vist_0}). 
In the recipe example, \model avoids major semantic errors and maintains consistent object shapes across steps; for instance, while StoryDiffusion and 1P1S show inconsistencies in the shape of the tins between steps 3 and 4, \model preserves them accurately.
In the VIST example, \model maintains both semantic alignment and visual coherence throughout the sequence, accurately depicting two friends and preserving consistent colors and lighting across all steps. In contrast, baselines such as 1P1S introduce a temporal inconsistency, shifting from sunset to night and back to sunset, while StoryDiffusion misinterprets the step text by showing three individuals instead of two friends.

\section{Limitations}
\label{sec:limitations}
Exploring the latent denoising space with \model requires Diffusion Models that provide access to fine-grain latents across denoising iterations, where \model still has the opportunity to steer the reverse diffusion process.

The theoretical complexity of BeamDiffusion is linear with the number of beams that we configure the model to explore. The drawback of this strategy is that exploring the space of hypothesis increases the computational complexity of the process, discussed in Appendix~\ref{appendix:challenges}.
However, this provides a solid ground to explore exciting new ideas where stronger constraints can be enforced in the explored beams.

\section{Conclusions}
In this paper, we proposed a novel method to explore the latent diffusion space with a beam search strategy, to generate image sequences.
\model is a theoretically sound framework, drawing on the strengths decoding techniques~\cite{shi2024thoroughexaminationdecodingmethods}, with  clear experimental gains over alternative methods, as we demonstrated in generating image sequences with enhanced coherence and consistency.

Through both human and automatic evaluation, we show that beam search performs effectively in generating high-quality and coherent image sequences. The insights drawn from these evaluations reveal that beam search aligns closely with textual instructions while ensuring visual continuity.
Most methods can align images with text, however, we demonstrated that \textit{sequence decoding methods perform naturally better at aligning their output with the input sequence}, and \model in particular achieves an optimal balance between the sequence of text instructions and the visual coherence of the generated images.
Furthermore, our results underscore the significance of the non-linear selection process in beam search, highlighting how prioritizing certain steps and latents can optimize the quality of the generation process.
This work shows that the latent diffusion space can be steered, offering more reliable and factual outputs. As future work, we plan to explore constraint decoding method that assess the likelihood of each denoise latent with respect to the target prompt. Solving this challenge, will improve the correlation between all elements in a sequence of scenes.

{
    \small
    \bibliographystyle{ieeenat_fullname}
    \bibliography{main}
}

\appendix

\clearpage
\section{Contextual Scene Descriptions}
\label{sec:contextual_scene_descriptions}

In many cases, step descriptions alone may lack context as precise visual prompts. These step descriptions are often interdependent, relying on prior steps for contextual understanding, or may miss visual details, e.g. \textit{"Put the ice cream in freezer for 1 hour"}, or actions that involve multiple tasks, such as \textit{"Slice the tomatoes, grate the cheese, and stir the sauce"}.
To address this challenge, we leverage Gemini~\cite{gemini}, specifically the Gemini 1.5 Flash model, to refine step descriptions into visually detailed prompts while considering sequence context. Specifically, Gemini produces a contextualized caption $c_j$ based on the current step description $s_j$ and all preceding steps as follows:
\begin{equation}
\textstyle 
c_j = \phi(s_j | \{s_1, s_2, \dots, s_{j-1}\})
\end{equation}
Following this approach we can improve image generation prompts by maintaining consistency and preserving the dependencies between sequential steps.

\section{Best Beam Search Configuration}
\label{appendix:beam_config}
The hyperparameters beam width ($w$ in Eq.~\ref{eqn:beam_width}) and steps back ($m$ in Eq.~\ref{eqn:latents_set}), play a crucial role in the performance of \model.  In this section, we report an exaustive evaluation on how different beam widths (2, 3, 4, and 6) and steps back (2, 3, and 4) improve the latent selection and the overall quality of the generation. A more detailed discussion on the impact of the latents selection is provided in Appendix~\ref{appendix:latents}. 

We first perform human annotations on a set of sequences to identify the optimal beam search configuration, see Appendix~\ref{appendix:annotations} for details. To determine the best configuration, annotations were used to discard non-optimal configurations. 
The results in Table~\ref{tab:human_best_config} showed that the best performing configuration was 2 steps back with a beam width of 4, achieving a selection rate of 25\%. Other configurations, such as 3 steps back with a beam width of 2 and 4 steps back with a beam width of 2, also showed a strong performance (16.67\%), but none outperformed the 2 steps back and beam width of 4 configuration. The configurations with beam width 6 performed the worst in all the settings tested.

To verify the reliability of our findings, the LLM-as-a-judge was applied to six tasks, yielding an 85\% Fleiss' Kappa agreement~\cite{agreement}, reflecting near-perfect alignment. 
The extended LLM-as-a-judge evaluation (Table~\ref{tab:gemini_best_config}) confirmed that the optimal configuration, 2 steps back with a beam width of 4, remained best, reaching a 48.57\% selection rate compared to 33.33\% in the initial six-task evaluation.
Further, the configuration with 3 steps back and a beam width of 2 still performed well, yielding a 28.57\% selection rate, confirming its robustness as an alternative. However, larger beam widths, such as beam width 6, continued to underperform, reinforcing that, in this task, a higher beam width does not always improve performance.

\begin{table*}[t]
    \centering
    \begin{minipage}{0.45\textwidth}
        \centering
        \caption{Influence of different hyperparameter values according to human annotations.}
        \label{tab:human_best_config}
        \begin{tabular}{c c c c c}
            \toprule
            \multirowcell{2}{\textbf{Steps}\\ \textbf{Back}} & \multicolumn{4}{c}{\textbf{Beam Width}} \\
            \cmidrule{2-5}
            & \textbf{2} & \textbf{3} & \textbf{4} & \textbf{6} \\
            \midrule
            \textbf{2} & 8.33 & 8.33 & \textbf{25.00} & 8.33\\ 
            \textbf{3} & \underline{16.67} & - & - & - \\
            \textbf{4} & \underline{16.67} & 8.33 & 8.33 & - \\
            \bottomrule
        \end{tabular}
    \end{minipage}
    \hspace{0.04\textwidth}
    \begin{minipage}{0.45\textwidth}
    \centering
        \caption{Influence of different hyperparameter values according to LLM-as-a-judge annotations.}
        \label{tab:gemini_best_config}
        \begin{tabular}{c c c c c}
            \toprule
            \multirowcell{2}{\textbf{Steps}\\ \textbf{Back}} & \multicolumn{4}{c}{\textbf{Beam Width}} \\
            \cmidrule{2-5}
            & \textbf{2} & \textbf{3} & \textbf{4} & \textbf{6} \\
            \midrule
            \textbf{2} & 5.71 & 2.86 & \textbf{48.57} & 5.71 \\
            \textbf{3} & \underline{28.57} & - & 2.86 & - \\
            \textbf{4} & - & - & 2.86 & - \\
            \bottomrule
        \end{tabular}
    \end{minipage}
\end{table*}

\section{Impact of Latent Selection on Beam Search Performance}
\label{appendix:latents}

Latent selection plays a crucial role in determining the quality and coherence of image sequences generated through beam search. By analyzing how different latent groups contribute to generation performance, we can better understand their impact on semantic alignment and visual coherence.

This aligns with recent work on interpreting diffusion models. For example, \citet{PontTuset_eccv2020} ground textual descriptions in image regions, while \citet{tang-etal-2023-daam} and \citet{dewan2024diffusionpidinterpretingdiffusionpartial} explore how textual prompts influence generation through attention analysis and partial information decomposition. In contrast, our method contributes to interpretability by examining how latent space traversal, guided by beam search, affects the generation of coherent, semantically aligned image sequences.

To systematically investigate this, we evaluate several sequences, each consisting of four steps, allowing us to assess the effects of latent selection across a diverse set of sequences. Using both automatic evaluation and human annotations, we analyze different latent configurations to determine which groups contribute most to generating coherent and semantically meaningful sequences.

\subsection{Impact of Latent Groups on Performance}


Following the initial analysis of beam search configurations (Appendix~\ref{appendix:beam_config}), where we used latent indexes 0, 1, 2, and 3, we further investigated the impact of using different latent configurations on the generated sequences. While latents 0, 1, 2, and 3 show promising results, latents 1, 3, 5, and 7 exhibit significantly poorer performance. In particular, beam search configurations for these underperforming latents result in very limited selection rates, indicating they are rarely chosen in both human and LLM-as-a-judge annotations. For human annotations across 6 tasks, the only configuration that yielded any selection was the one with 4 steps back and a beam width of 3, achieving a selection rate of 8.33\%. All other configurations resulted in 0\% selection. In contrast, none of the configurations showed any non-zero selection rate under LLM-as-a-judge annotations, reinforcing the lack of preference for these latents.

\begin{figure}[h!]
\centering
 \includegraphics[width=1\columnwidth]{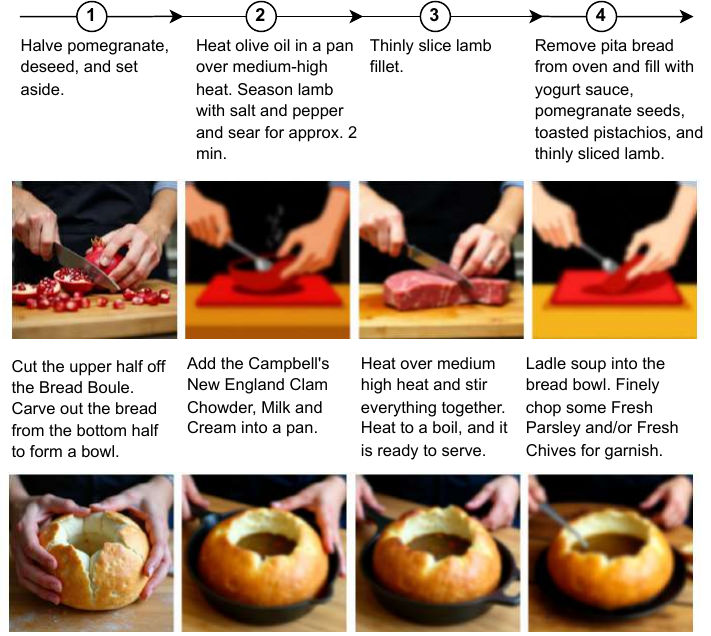} 
        \caption{Impact of using late-index latents configurations on sequences generation.}
        \label{fig:latents_example}
\end{figure}

Similarly, for LLM-as-a-judge annotations evaluated on the same 6 sequences, the configurations do not yield any notable results, with no selection rates. When the evaluation is extended to 35 sequences, the performance for latents 1, 3, 5, and 7 remains sparse, with a single configuration (2 steps back, beam width of 3) being selected, with a selection rate of 2.86\%. Figure~\ref{fig:latents_example} visually demonstrates the poor results, where the images appear nearly identical, further supporting the findings of limited diversity and low performance, which aligns with the findings of \cite{CoSeD, bordalo24}.

\subsection{Latent Selection on Image Sequence Generation}
\label{appendix:latent_selection}
\begin{figure}[h!]
    \centering
    \includegraphics[width=1\columnwidth]{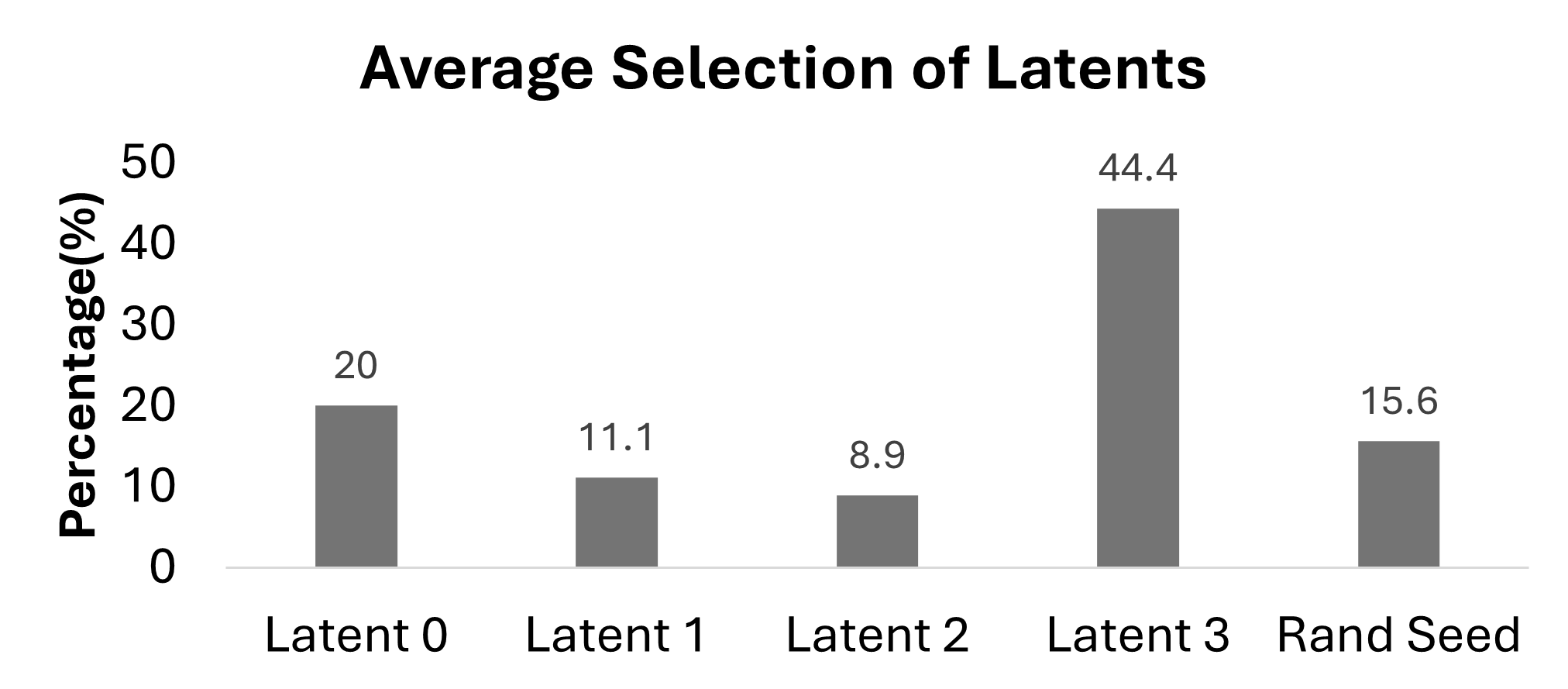} 
    \captionof{figure}{The average number of times that the model selected a given \textbf{latent} to generate the next image.}
    \label{fig:latent_selection}
\end{figure}

Following our evaluation of the impact of latent groups on performance, we now examine the role of individual latents within the higher-performing group (Latents 0-3). As shown in Figure~\ref{fig:latent_selection}, Latent~3 is selected most frequently (44.4\%), followed by Latent~0 (20\%), while Latents~1 and~2 are used less often (11.1\% and 8.9\%, respectively). This uneven distribution suggests that Latents 0 and 3 contribute disproportionately to the generation of sequences with better text alignment.
One possible explanation is that these latents encode more semantically relevant or visually discriminative features that align well with the text prompts. Their frequent selection by beam search indicates that these latents are particularly effective at resolving ambiguities or reinforcing important visual cues during sequence generation. The lower usage of Latents 1 and 2 may reflect their tendency to encode less salient or redundant information, which contributes less to text-conditioned coherence.
Interestingly, the Rand seed, which injects stochasticity into the generation process, accounts for 15.6\% of selections. This suggests that while beam search primarily exploits strong latents like 0 and 3, it also values diversity. The inclusion of randomized elements allows the search to escape local optima and sample from alternative yet plausible paths, enhancing the robustness of the final sequence.

\subsection{Effect of Cumulative Latent Access}
\label{appendix:cumulative_latents}
To investigate how access to latents from previous steps influences the quality of generation, we conducted an ablation study in which we constrain the set of available latents throughout the generation of the sequence. Specifically, we evaluate performance when the model is limited to latents from only the first step, then from the first two steps, and so on, up to including latents from all previous steps (i.e., full cumulative access).

\begin{figure}[t]
    \centering
    \includegraphics[width=1\columnwidth]{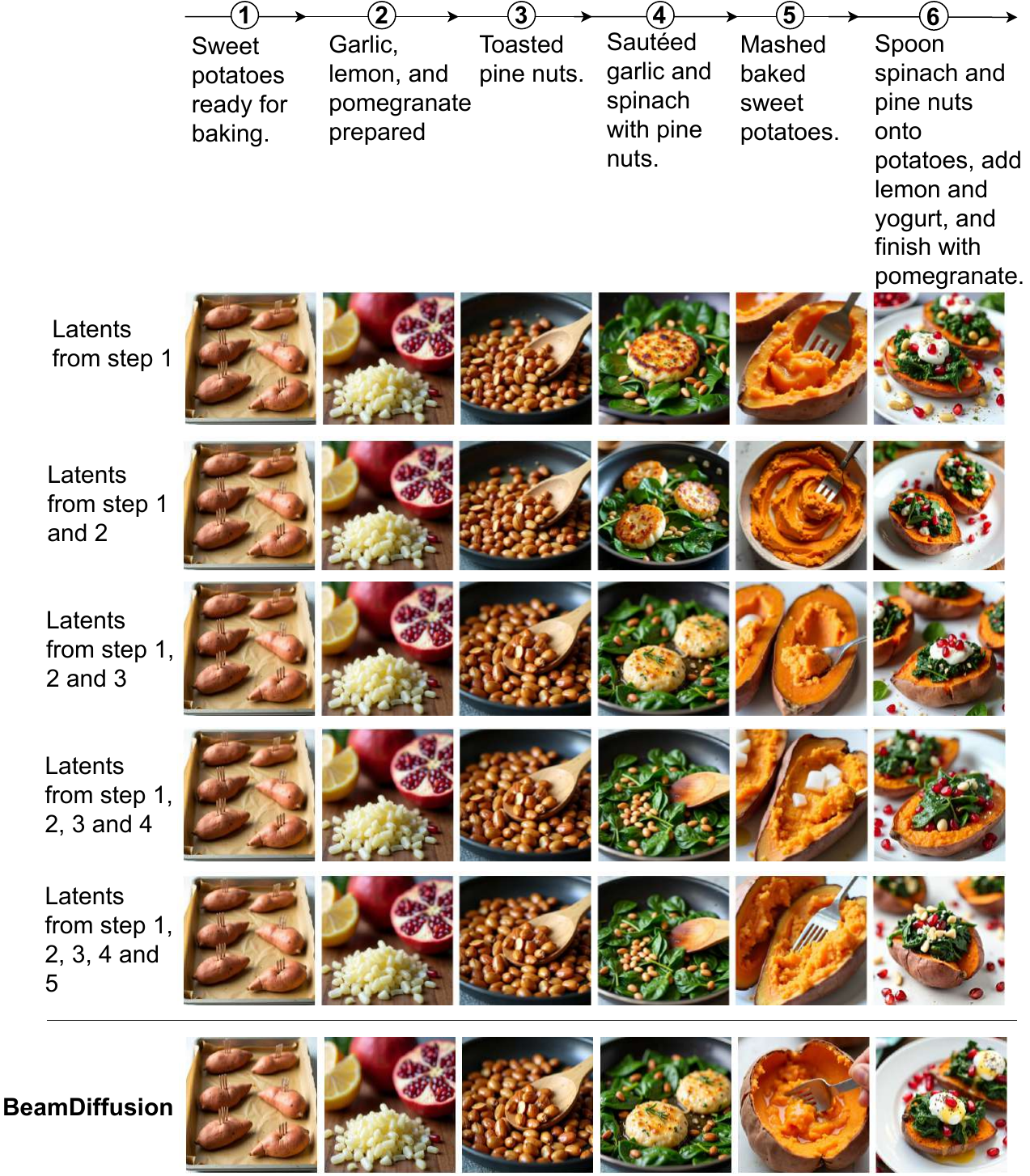} 
    \caption{Visual comparison of sequences generated with cumulative latent access. Each row shows a sequence generated using latents from only the first step, the first two steps, and so on, up to all previous steps. The bottom row shows the result from \model, which uses latents from only the two preceding steps.}
    \label{fig:cumulative_latents}
\end{figure}

As shown in Figure~\ref{fig:cumulative_latents}, including latents from only the first step makes step 4 appear very closed up. Limiting the model to latents from the first two steps makes it more difficult to follow the step text while maintaining coherence, which is particularly evident in step 5. Gradually expanding the latent set to include more steps improves sequence quality, especially when moving from two to three steps. However, including latents from all previous steps does not yield further improvements in coherence, suggesting diminishing returns with increasing temporal access.

In contrast, \model uses latents from only the two most recent steps and achieves results that are comparable, or even superior, to the full cumulative case. The reason is that sequence generation naturally forms a latent chain: to generate a new image, the model conditions on a latent from the previous step, which has already absorbed information from earlier generations. The resulting latent is then updated with the new step, effectively carrying forward both old and new information (see Figure~\ref{fig:img_gen} for an illustration of this chain effect). In this way, the most recent latents act as a compressed summary of the entire history, making access to all earlier latents redundant. By always using only the last two latents, \model maintains a global summary of the sequence while preserving efficiency.

Overall, this experiment highlights that cumulative latent access enhances generation quality up to a point, but full historical memory is unnecessary. Strategies such as \model, which leverage a limited yet informative temporal window, can produce high-quality sequences efficiently. These findings also underscore the importance of the non-linear sequence denoising strategy introduced in Section~\ref{sec:non_linear_sequence_denoising}, which enables more targeted and effective reuse of past latents.

\subsection{Automatic and Qualitative Evaluation of Decoding Strategies}
\label{appendix:qual_decoding}

\begin{table*}[h]
\centering
\caption{Comparison of different decoding methods across embedding-based sequence quality metrics. CLIP-I, DINO-I, CLIP-T, CLIP*, and DINO* are reported as win percentages, while Goal Faithfulness, Step Faithfulness, and Cross-Image Consistency are reported as absolute scores.}
\label{tab:merged_comparison}
\resizebox{\textwidth}{!}{%
\begin{tabular}{@{}lccccc|cccc@{}}
\toprule
\textbf{Method} & \textbf{CLIP-I} & \textbf{DINO-I} & \textbf{CLIP-T} & \textbf{CLIP*} & \textbf{DINO*} & 
\makecell{\textbf{Goal Faith.}$\uparrow$ } &
\makecell{\textbf{Step Faith.}$\uparrow$} &
\makecell{\textbf{Cross-Image Cons.}$\uparrow$} \\ 
\midrule
Greedy/CoSeD$_{len=1}$  & \underline{30.00} & \underline{30.00} & \underline{35.00} & 25.00 & \underline{30.00} & \underline{0.76} & \underline{0.90} & \textbf{0.62}\\
Nucleus Sampling & 25.00 & \textbf{40.00} & \textbf{40.00} & \underline{35.00} & \textbf{35.00}  & 0.72             & 0.80             & 0.60 \\
BeamDiffusion & \textbf{45.00} & \underline{30.00} & 25.00 & \textbf{40.00} & \textbf{35.00} & \textbf{0.82} & \textbf{0.92} & \underline{0.61} \\
\bottomrule
\end{tabular}%
}
\end{table*}

\begin{figure}[h!]
    \centering
    \includegraphics[width=1\columnwidth]{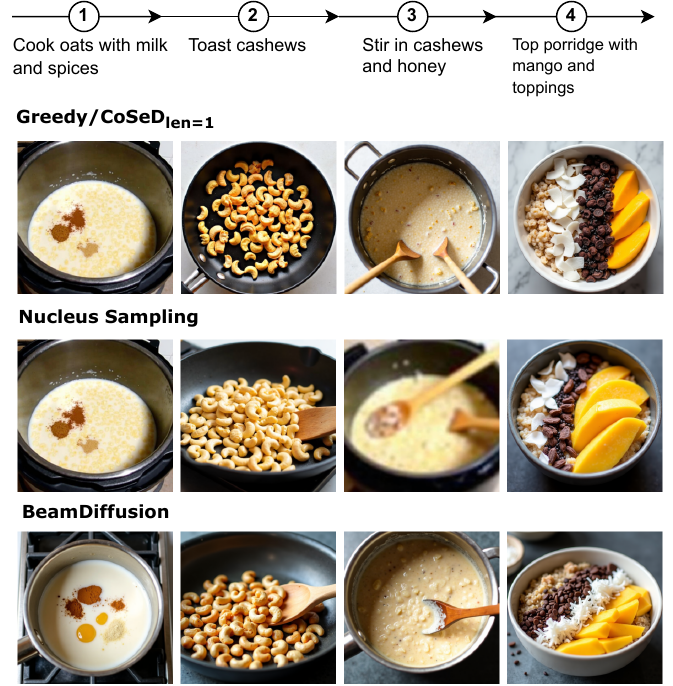}
    \caption{Illustrative comparison of decoding approaches. \model prevents cascading inconsistencies across steps while preserving semantic and visual consistency.}
    \label{fig:recipe31_dec}
\end{figure}

\paragraph{Automatic evaluation.}  
Table~\ref{tab:merged_comparison} reports automatic comparisons across similarity metrics and faithfulness/consistency measures. 
The combined scores CLIP* and DINO* offer a more holistic signal than their individual components. 
Here, \model achieves the best CLIP* score (40.00) and matches the highest DINO* (30.00), indicating strong performance across both text–image and image–image similarity. 
By contrast, Greedy/CoSeD lags significantly on CLIP* (25.00) and only reaches 30.00 on DINO*, while Nucleus Sampling improves slightly on CLIP* (35.00) and DINO* (35.00) but at the expense of weaker instruction alignment and coherence. 

\model also obtains the strongest scores on goal faithfulness (0.82) and step faithfulness (0.92), showing that its generations remain closely aligned with both the overall task and individual step instructions. 
Greedy/CoSeD achieves reasonably high step faithfulness (0.90) and the best cross-image consistency (0.62), but falls behind in goal alignment. 
Nucleus Sampling, in contrast, is weaker across all three measures (0.72, 0.80, 0.60). 
Taken together, the similarity metrics and the faithfulness/consistency measures highlight that \model avoids the sharp trade-offs of the other decoding methods, achieving a favorable balance between multimodal similarity, instruction following, and visual coherence.

\begin{figure}[h!]
    \centering
    \includegraphics[width=1\columnwidth]{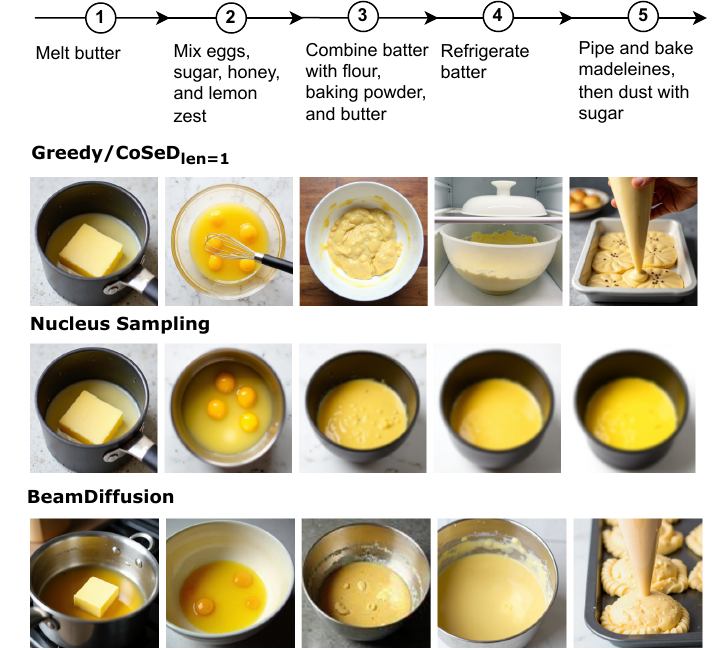}
    \caption{Illustrative comparison of decoding approaches. \model prevents cascading inconsistencies across steps while preserving semantic and visual consistency.}
    \label{fig:recipe3_dec}
\end{figure}

\paragraph{Qualitative evaluation.}  
Figures~\ref{fig:recipe31_dec} and \ref{fig:recipe3_dec} illustrate characteristic error modes of baseline strategies. Nucleus Sampling often suffers from blur propagation, where low-frequency artifacts introduced in one step persist across subsequent generations, degrading the sequence. Greedy decoding avoids blur but instead produces visual inconsistencies, with objects abruptly shifting style or geometry across steps (e.g., the changing appearance of the pan between step 1 and step 3, in Figure~\ref{fig:recipe31_dec}).  

In contrast, \model avoids both pitfalls. By dynamically selecting informative latent trajectories, it prevents error accumulation while preserving smooth semantic and visual transitions. The generated sequences remain sharper, semantically faithful, and more coherent over time. This demonstrates that decoding is not merely a technical detail but a decisive factor in determining the fidelity and consistency of long-horizon generations.

\paragraph{Summary.}  
Overall, both automatic and qualitative results show that \model delivers a principled improvement over standard strategies. By explicitly managing latent trajectories, it achieves robustness against error propagation and inconsistency while preserving semantic faithfulness and cross-step coherence.

\section{Minimizing Decoding Risk}
\label{appendix:minimizing_decoding_risks}
The generation process using diffusion models begins with a seed that determines the initial conditions of the model. The choice of this seed significantly affects the outcome, as it influences the path the model takes during generation. Using multiple random seeds in the first step ensures a diverse exploration of possible outputs, avoiding premature narrowing of options into a single path based on that seed. This limits the diversity of the results and increases the chance of missing better options.

\paragraph{Greedy Decoding and Nucleus Sampling.} 
In both greedy decoding and nucleus sampling, we use the same approach in the first step. Multiple random seeds are employed to generate a set of candidate images. Each candidate is then evaluated using the CLIP score mechanism, which compares the generated images with the step text. The image that best aligns with the text, according to the CLIP score, is selected to continue the process. This method ensures that both greedy decoding and nucleus sampling benefit from multiple diverse starting points, increasing the quality and variety of the generated outputs.

\paragraph{\model.}
In \model, relying on a single random seed in the first step effectively makes the process a greedy selection, as it locks the generation into a specific path. This limits diversity and reduces the chances of discovering higher-quality outputs. By using multiple random seeds, each seed generates a different decoding trajectory, helping to broaden the search space and increasing the likelihood of finding better solutions.
Using multiple random seeds in the initial step enhances the exploration of the latent space, ensuring better and more diverse results across different decoding strategies.

\paragraph{Random Seeds for mid-sequence Generation.}The sequence of scenes \( S = \{s_1, s_2, \dots, s_L\} \) may have one scene in the middle of the sequence that should be independent. To address this, for steps $s_{j>1}$, we use a combination of random seeds and past latent vectors to maintain a balance between diversity and coherence, which is particularly important for such scenarios. 

\begin{figure}[t]
    \centering
    \includegraphics[width=1\columnwidth]{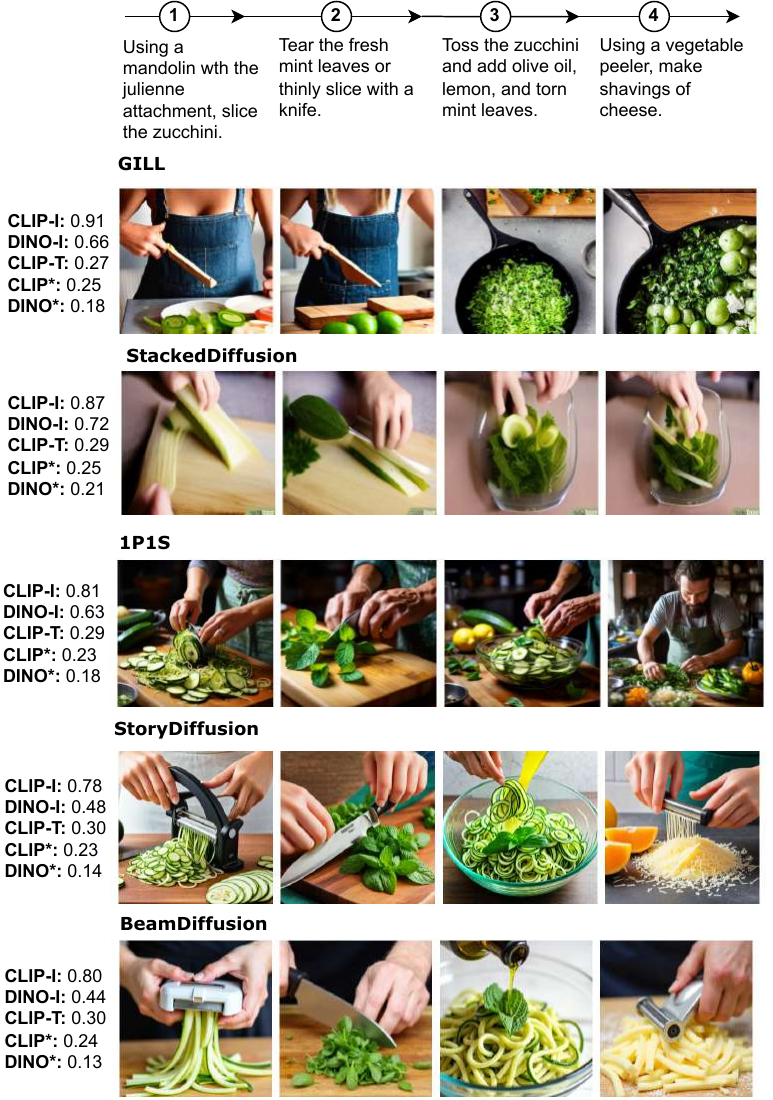} 
    \caption{Illustration of the limitation of CLIP-I: despite a high CLIP-I score, the result is not semantically correct or visually faithful. The example compares several automatic metrics (CLIP-T, CLIP-I, DINO-I, CLIP*, DINO*), highlighting discrepancies and failure cases in sequence evaluation.}
    \label{fig:metric_limitation}
\end{figure}

\section{Metrics and Annotations}
\label{appendix:annotations}
\paragraph{Automatic Metrics.}
In image sequence generation, CLIP similarity scores between visually similar frames often exceed 0.85, even for near-duplicate content. Wang et al.~\cite{WangZSC24} highlight that such high CLIP-I scores can indicate semantic duplication, which poses a challenge when diversity across frames is desired. While CLIP-I and DINO-I are common proxies for semantic similarity, rewarding near-duplicates undermines meaningful visual progression.
Conversely, poor alignment between images and their textual prompts is reflected in low CLIP-T scores. Lin et al.~\cite{LinExplore} address this by clipping CLIP-T values below 0.1, effectively filtering out weak text-image matches.

To balance these factors, we set CLIP-I scores above 0.9 and DINO-I scores above 0.85 to zero, penalizing visual redundancy. Similarly, we zero out CLIP-T scores below 0.1 to discourage poor alignment with the prompt. This clipping strategy encourages the generation of sequences that are both semantically relevant and visually diverse.
In Figure~\ref{fig:metric_limitation}, we present the automatic metric scores for a sequence across all methods. The results illustrate that very high CLIP-I values can lead to sequences that fail to follow the textual steps accurately, as observed in the case of GILL. Despite achieving strong visual similarity, the sequence diverges from the intended narrative, highlighting a key limitation of relying solely on CLIP-I . This example reinforces the importance of our clipping strategy: without penalizing excessively high CLIP-I values, the generation process may favor near-duplicates over meaningful visual progression, ultimately undermining semantic fidelity to the prompt.

Even after applying this clipping, we observe that automatic metrics remain imperfectly aligned with human and LLM-as-a-judge evaluations, underscoring the ongoing challenge of developing reliable metrics for evaluating sequences of images.

\paragraph{Human annotations.}To facilitate the selection of the best beam search configuration, we developed a custom website specifically for human annotation. We found that existing tools lacked the flexibility necessary for our specific use case, especially when evaluating multiple configurations simultaneously. To select the best beam search configuration, 5 annotators are presented with 24 different sequences, each corresponding to a unique beam search configuration.

In total, the 24 configurations were generated by combining three key factors:
\begin{itemize}
    \item \textbf{Steps Back} ($m$ in Eq.~\ref{eqn:latents_set}): This refers to how many previous steps the model uses from its latent representations to explore new candidate outputs. We evaluated three different settings for the number of steps back: 2, 3, and 4, which correspond to using latents from the last two, three, or four images, respectively.
    \item \textbf{Latents Indexes}: We tested two sets of latent configurations, {1, 3, 5, 7} and {0, 1, 2, 3}, by selecting specific indexes from the latent representations generated in previous stages. This was done to explore how varying levels of latent detail impact the quality of the generated sequences.
    \item \textbf{Beam Width} ($w$ in Eq.~\ref{eqn:beam_width}): We explored four different beam widths, 2, 3, 4 and 6, which determine how many beams are retained at each step during the search process.
\end{itemize}

\begin{figure}[t]
    \centering
    \includegraphics[clip, trim= 0 0 0 40, width=1\columnwidth]{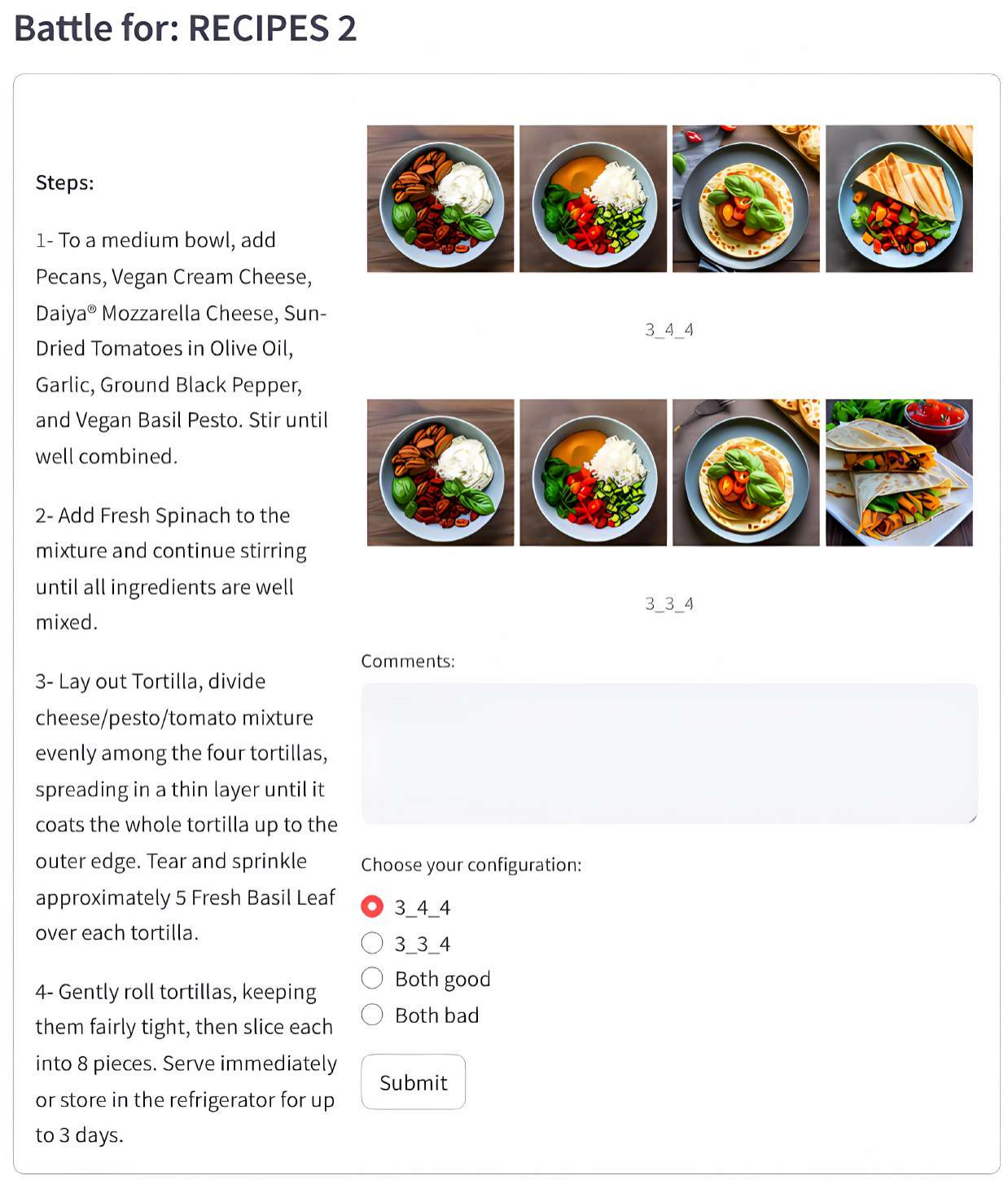}
    \caption{Example of the human annotation elimination round for selecting the best beam search configuration.}
    \label{fig:battle_example}
\end{figure}

Evaluating all of these at once proved to be exhausting and difficult to manage.
To simplify the evaluation process, we adopted a elimination annotation method. In this approach, 5 annotators compare two configurations at a time and select the one with better overall coherence. If neither configuration stands out, annotators can select \textit{"both good"} or \textit{"both bad"} (Figure~\ref{fig:battle_example}), in which case the computationally cheaper option advances to the next round. This process is repeated until we identify the best configuration. This method focuses on direct comparisons, making it easier to pinpoint the most effective beam search settings while also considering computational efficiency.

\begin{figure}[t]
    \centering
    \begin{subfigure}[t]{0.48\columnwidth}
        \includegraphics[clip, trim=120 355 310 115, width=\linewidth]{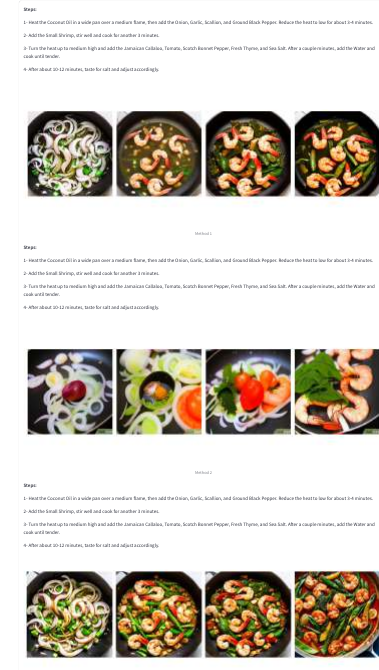}
    \end{subfigure}
    \hfill
    \begin{subfigure}[t]{0.48\columnwidth}
        \includegraphics[clip, trim=120 30 310 445, width=\linewidth]{figures/method_selection_example.pdf}
    \end{subfigure}
    \caption{Example of a human annotation to choose the best method.}
    \label{fig:method_selection_example}
\end{figure}

To evaluate all models performance, annotators are presented with five generated sequences side by side.
Figure~\ref{fig:method_selection_example} illustrates an example of this comparison process, where the annotators assess the sequences to determine the most effective approach.

\paragraph{LLM-as-a-judge Evaluation.}
The LLM-as-a-judge selection process for annotating the best beam search configuration and generation method closely mirrored that of human annotators, with one key adjustment: the model tended to default to \textit{"both good"} or \textit{"both bad"} when such options were available. To encourage more decisive outcomes, we removed these options from the prompt. We also refined our comparison methodology. Presenting all five sequences simultaneously initially led to biased evaluations, as the model disproportionately favored the last two. To address this, we adopted a pairwise evaluation strategy, comparing two sequences at a time. This adjustment yielded more balanced and reliable assessments across all comparisons.

Figure~\ref{fig:gemini_prompt} presents the LLM-as-a-judge prompt for evaluating image sequences on visual and semantic consistency.
\begin{figure}[t!]
\centering
\begin{tcolorbox}[fontupper=\scriptsize, fontlower=\scriptsize]
            \textbf{LLM-as-a-judge Prompt:} \\
            Evaluate two image sequences using the following criteria:
            \begin{itemize}
                \item \textbf{Visual Consistency}: Objects, ingredients, backgrounds, colors, and styles should remain consistent across frames. Frames should transition smoothly, resembling a video-like progression with no sudden changes or missing elements.
                \item \textbf{Semantic Consistency}: Images must logically follow the described steps, clearly and accurately.
            \end{itemize}
            
            \textbf{Steps of the process:} \{steps\}
            
            After evaluating both sequences, select the one with the strongest overall consistency (visual and semantic). Pay close attention to even minimal differences, as the sequences are very similar. If both sequences are equally good or bad, choose the one that is slightly better. Choose exactly one of the following options:\\
            \tab - "First sequence"\\
            \tab - "Second sequence"
\end{tcolorbox}
\caption{LLM-as-a-judge prompt used to select the best beam search configuration and best method.}
\label{fig:gemini_prompt}
\end{figure}

\section{Datasets}
\label{appendix:datasets}
\paragraph{Recipes.} Each manual task in the Recipes datasets includes a title, a description, a list of ingredients, resources and tools, and a sequence of step-by-step instructions, which may or may not be illustrated. The Recipes dataset comprises approximately 1,400 tasks, with an average of 4.9 steps per task, totaling 6,860 individual steps. Most tasks feature an image for each step. 

\paragraph{VIST.}We used the Descriptions of Images-in-Isolation (DII) annotations, as they allow for the generation of more precise and informative visual prompts. These annotations provide standalone descriptions of images, making it easier to capture their key visual elements without relying on surrounding context. The dataset includes 50 000 stories, offering a diverse range of visual and textual information. To ensure the most accurate and relevant annotation for each image, we leveraged CLIP. By using CLIP, we were able to select the annotation that best represented the image, improving the quality of our generated prompts and enhancing the overall alignment between text and visuals.

\section{Challenges}
\label{appendix:challenges}
In the development and evaluation of the \model method, we encountered several challenges that required careful consideration and innovation. These challenges spanned multiple aspects of the model, ranging from the difficulty of generating accurate and concise captions, to identifying suitable domains for testing, and addressing the computational complexity of the generation process. While we were able to propose solutions for some of these challenges, others remain open and present ongoing areas of research. In this section, we outline the key challenges we faced and the strategies we employed, while acknowledging that not all of them have been fully addressed.
\paragraph{Captions.}
One of the challenges encountered in this process is generating captions that accurately describe the steps involved in the sequence. Often, a single step may consist of multiple sub-steps, each contributing to the overall generation. This makes it difficult to write a single concise caption that captures the full scope of each step. The challenge lies in ensuring that the caption not only reflects the specific action or transformation occurring at each step but also conveys the progression and complexity of the entire sequence. Additionally, if we divide the steps into smaller sub-steps to better describe the process, the sequence becomes very extensive, making it even harder to maintain clarity and readability while ensuring comprehensive coverage of the entire generation process.

\paragraph{Domains.}
Identifying a domain that effectively highlights the advantages of the \model method, particularly the benefits of beam search, is still a challenge. While \model provides improvements in consistency and coherence, not all domains clearly showcase these advantages. Certain domains may lack the complexity or diversity needed to demonstrate the method's strengths, such as its adaptive latent selection or its ability to maintain alignment with contextual instructions. Finding a domain that not only aligns with the method's capabilities but also provides a clear and compelling demonstration of its improvements remains an ongoing challenge.

\paragraph{Complexity of the generation.}
The \model model generates multiple candidate images at each step by exploring a range of latents. We analyze the complexity of the model for a steps back of \(m\) steps and \(N \) latents explored. 
In the initial steps, the complexity is high because pruning begins only after the second step. For each beam, \(N\cdot m\) latents are explored, and the search space grows exponentially, resulting in significant computational cost in the early stages.
During the first two steps, the complexity grows with the number of candidates and latents, proportional to \( \mathcal{O}(N\cdot m)^j) \) for \( j \leq 2 \). Specifically, for $j=1$, the complexity is proportional to the number of explored random seeds. After the second step, pruning reduces the number of candidates to the most promising \(w\) beams, decreasing the complexity to \( \mathcal{O}(N\cdot m \cdot w)\).

To further reduce complexity, a heuristic based on text relevance can be used to predict the top $P$ most relevant past images. By restricting the search space to the latents of these top $P$ candidates, the model improves efficiency by prioritizing contextually relevant images, particularly in the early steps. While the initial three steps remain computationally demanding due to the expansive search space, pruning combined with this heuristic can significantly lower the computational cost by refining the candidate set.

\begin{figure}[t]
    \centering
    \begin{subfigure}[b]{1\columnwidth}
        \centering
        \includegraphics[width=\textwidth]{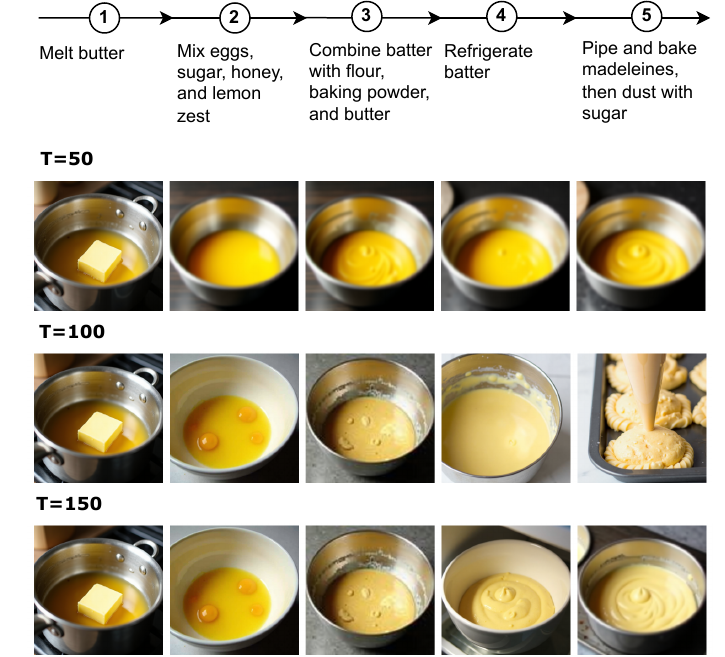}
        \caption{Example 1: 50-step generations exhibit blur propagation, while 100 steps reduce blur and sharpen frames. 150 steps show no substantial gain over 100.}
        \label{fig:denoising_it_example1}
    \end{subfigure}
    \vspace{0.3cm}
    
    \begin{subfigure}[b]{1\columnwidth}
        \centering
        \includegraphics[width=\textwidth]{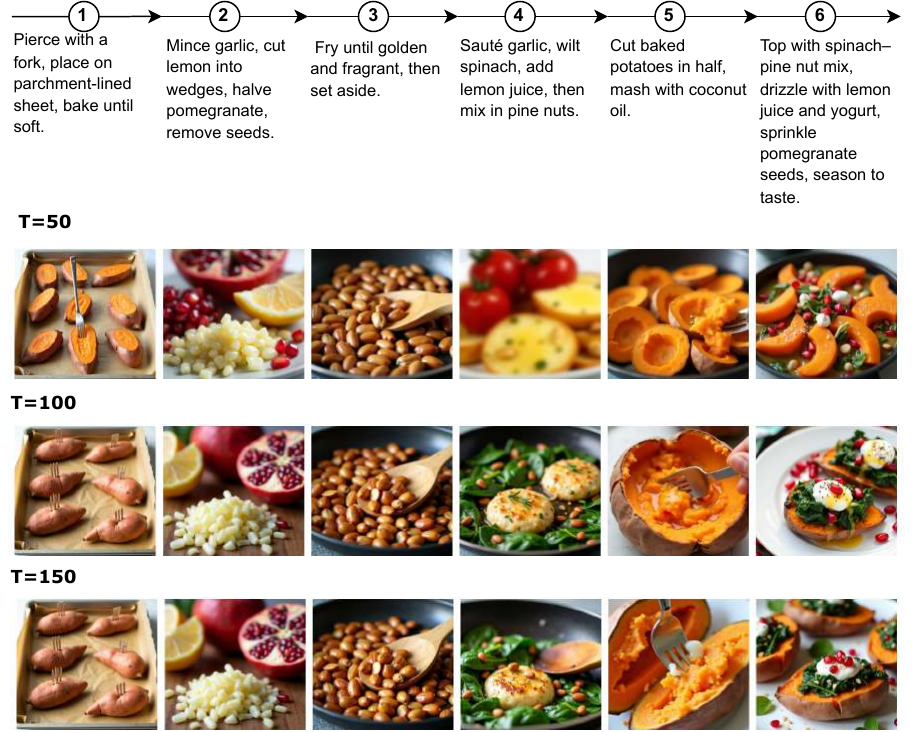}
        \caption{Example 2: 50 steps produce blurry, inconsistent frames, while 100 steps improve fidelity and coherence. 150 steps do not yield further noticeable improvements.}
        \label{fig:denoising_it_example2}
    \end{subfigure}
    
    \caption{Qualitative comparison of 50 vs. 100 denoising steps across two examples. Increasing steps to 150 does not produce substantial gains over 100, indicating diminishing returns.}
    \label{fig:denoising_it_comparison}
\end{figure}

\section{Implementation Details}
\label{appendix:implementation_details}

\paragraph{Base Models.} 
Our method is primarily implemented on top of FLUX.1-dev\cite{blackforestlabs_flux}, a diffusion transformer–based model released by Black Forest Labs. We extend its open-source implementation to support beam-based generation, latent trajectory tracking, and contrastive scoring mechanisms as described in the main text. Unless otherwise specified, we follow the default configuration of FLUX.1-dev with 100 denoising steps, chosen to balance generation quality and computational efficiency. While the default 50 steps generally produce globally coherent images, we observe that finer structures often remain underdeveloped, resulting in blur and reduced visual consistency across sequences. Increasing to 100 steps substantially sharpens details and improves sequence coherence, as illustrated in Figures~\ref{fig:denoising_it_example1} and~\ref{fig:denoising_it_example2}. Further increasing to 150 steps provides no additional gains, confirming that 100 steps is a reasonable trade-off. To ensure that the choice of decoding method is not dependent on a specific backbone, we also implement \model on Stable Diffusion v2.1\cite{rombach2022high}, a widely used U-Net–based latent diffusion model, using its standard parameters. This setup verifies that our conclusions regarding decoding strategies generalize across different generative architectures.

\paragraph{Latent Trajectory Reuse.} In our approach, maintaining coherence across sequential generations requires access to the intermediate latent representations produced during the generative process. Fortunately, the model pipeline inherently provides the full sequence of latent representations at each generation step. This enables us to easily extract and reuse specific latents as seed for new generations.

\paragraph{Classifier Training.}  
To train the contextual classifier, we employ a sequence-aware training approach that enables the model to capture dependencies between steps in the generation process. The classifier \(\varphi\) is primarily trained on the Recipes dataset, which provides rich sequential image-text pairs of cooking steps, allowing it to learn meaningful context transitions in a structured procedural domain. Additionally, a small amount of data from the VIST dataset is included during training to introduce limited variability in narrative structure. The classifier takes as input a candidate image \(I_l\), its textual description \(s_l\), and the previous sequence images and respective step text (\(B_{l-1}^{(b)}\)). It then scores \(I_l\) based on how well it fits within the evolving trajectory.

During training, we use labels \(l_{d,l}\) indicating whether \(I_l\) is the correct continuation in the sequence for task \(d\) and step \(l\). The model is trained to minimize the cross-entropy loss:
\begin{equation}
\textstyle
\argminB \sum_{d=1}^{D} \sum_{l=1}^{L} l_{d,l} \cdot \log(\varphi(I_l, s_l, B^{(b)}_{l-1})),
\end{equation}
This contrastive training encourages the model to assign high scores to contextually appropriate continuations while penalizing incoherent or semantically unrelated ones. As a result, the classifier learns to distinguish subtle differences in step-to-step coherence, both visually and textually.

\paragraph{Hardware.} All experiments were conducted on a NVIDIA A100 GPU with 40GB of memory.

\section{Broader Impacts}
\label{appendix:broader_impacts}
This work proposes a method to improve generative pretraining by incorporating subject-aware mutual information maximization, aiming to produce more contextually aligned and coherent sequences of images. The ability to generate visually consistent image sequences has positive applications in areas such as animation, visual storytelling, educational media, and accessibility tools (e.g., for generating illustrative content from text).
However, as with many generative models, there are potential negative societal impacts. Enhanced coherence and subject fidelity could be misused to create misleading or deceptive visual content, such as more convincing visual disinformation, propaganda, or synthetic media (e.g., deepfakes).

\section{Application Examples}
This section presents additional visual examples and qualitative results that complement the main paper. Images and other visual aids are essential tools that significantly improve comprehension across a variety of domains. Whether breaking down complex tasks, illustrating a sequence of events, or clarifying intricate concepts, visual representations provide clear, step-by-step guidance that enhances understanding~\cite{MAYER200285}. From technical procedures and instructional design to storytelling and educational content, multi-scene image sequences help readers grasp information more intuitively, transforming even the most complex or dense topics into something more accessible.
However, generating a sequence of images that not only aligns with each textual instruction but also maintains overall coherence remains a major challenge~\cite{lu2023multimodalproceduralplanningdual}. In this appendix, we include expanded examples to further illustrate how our model addresses this issue. These include additional visualizations of the beam search process and extended qualitative comparisons across domains such as recipes and visual storytelling. Together, these examples highlight the strengths and limitations of current approaches in achieving textual alignment and visual consistency in sequential image generation.

\subsection{Sequence Examples}
\label{sec:sequence_examples}
\begin{figure*}[htb!]
    \centering
    \begin{minipage}{.45\textwidth}
        \centering
        \includegraphics[width=1\columnwidth]{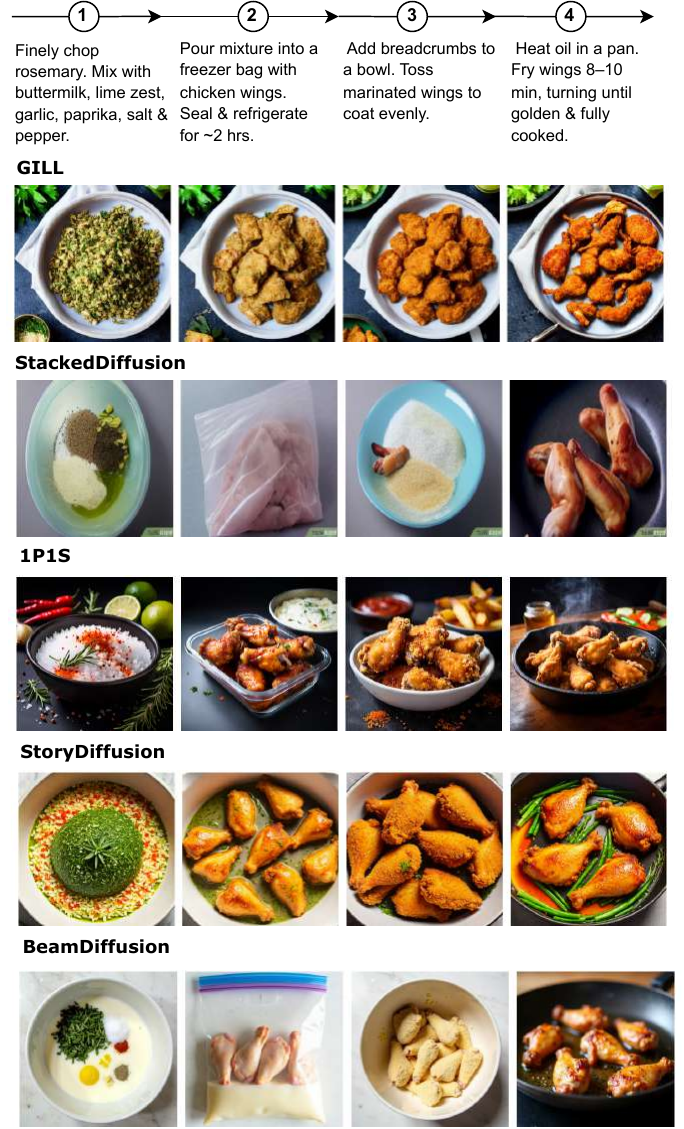} 
        \caption{Example of the Recipes domain task \textit{"Buttermilk chicken wings"}, with a sequence of 4 steps.}
        \label{fig:recipe_13}
    \end{minipage}%
    \hspace{5mm}
    \begin{minipage}{0.45\textwidth}
         \centering
        \includegraphics[width=1\columnwidth]{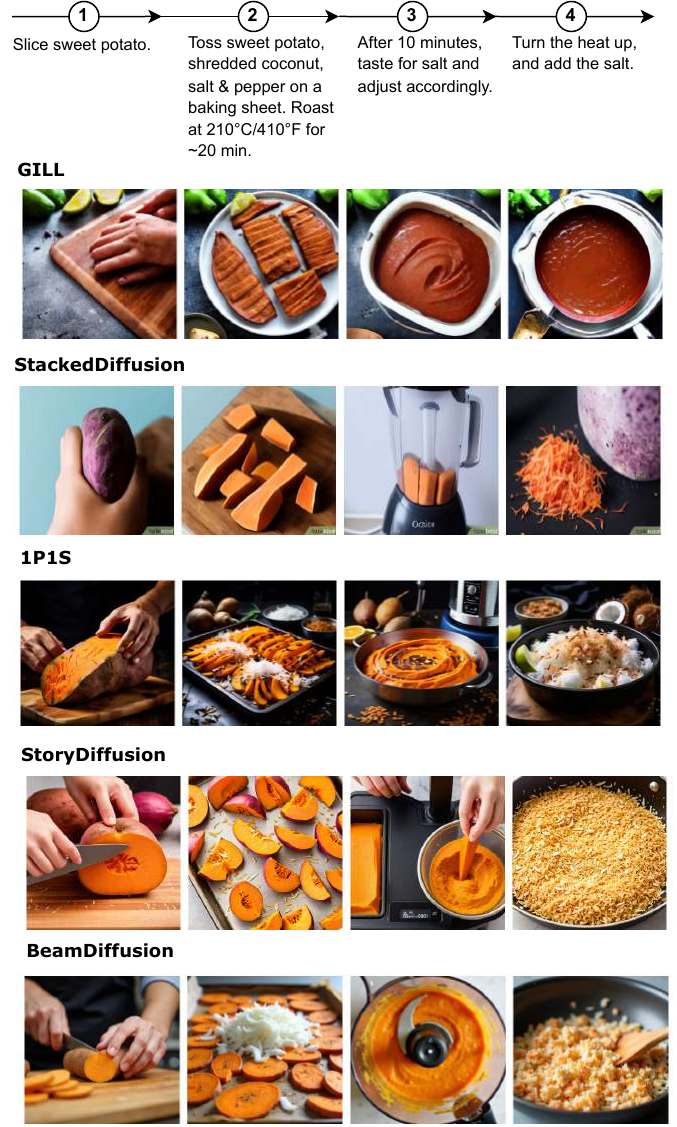} 
        \caption{Example of the Recipes domain task \textit{"Sweet potato and orange soup"}, with a sequence of 4 steps.}
        \label{fig:recipe_42}
    \end{minipage}
\end{figure*}

\begin{figure*}[htb!]
    \centering
    \begin{minipage}{.45\textwidth}
        \centering
        \includegraphics[clip, trim= 30 0 0 0, width=1\columnwidth]{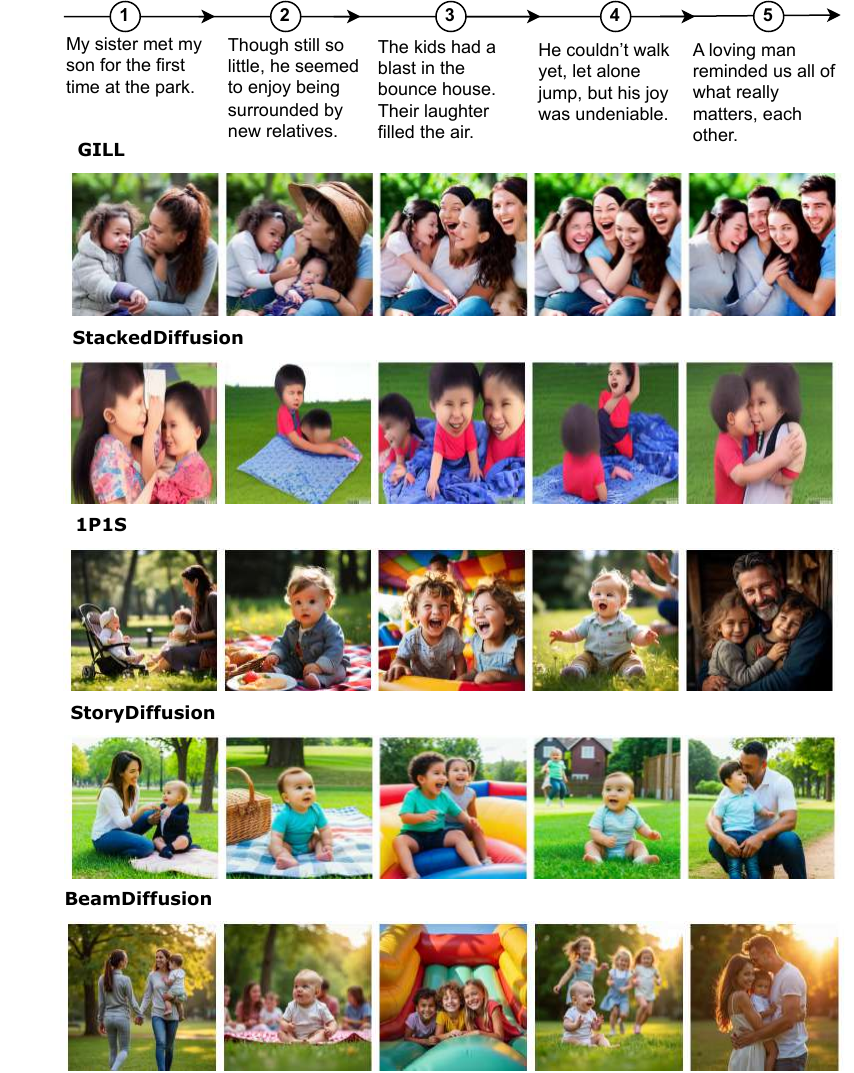} 
        \caption{Example of the VIST story domain \textit{"4th of July 2006"}, with a sequence of 5 steps.}
        \label{fig:vist_4}
    \end{minipage}%
    \hspace{5mm}
    \begin{minipage}{0.45\textwidth}
        \includegraphics[clip, trim= 30 0 0 0, width=1\columnwidth]{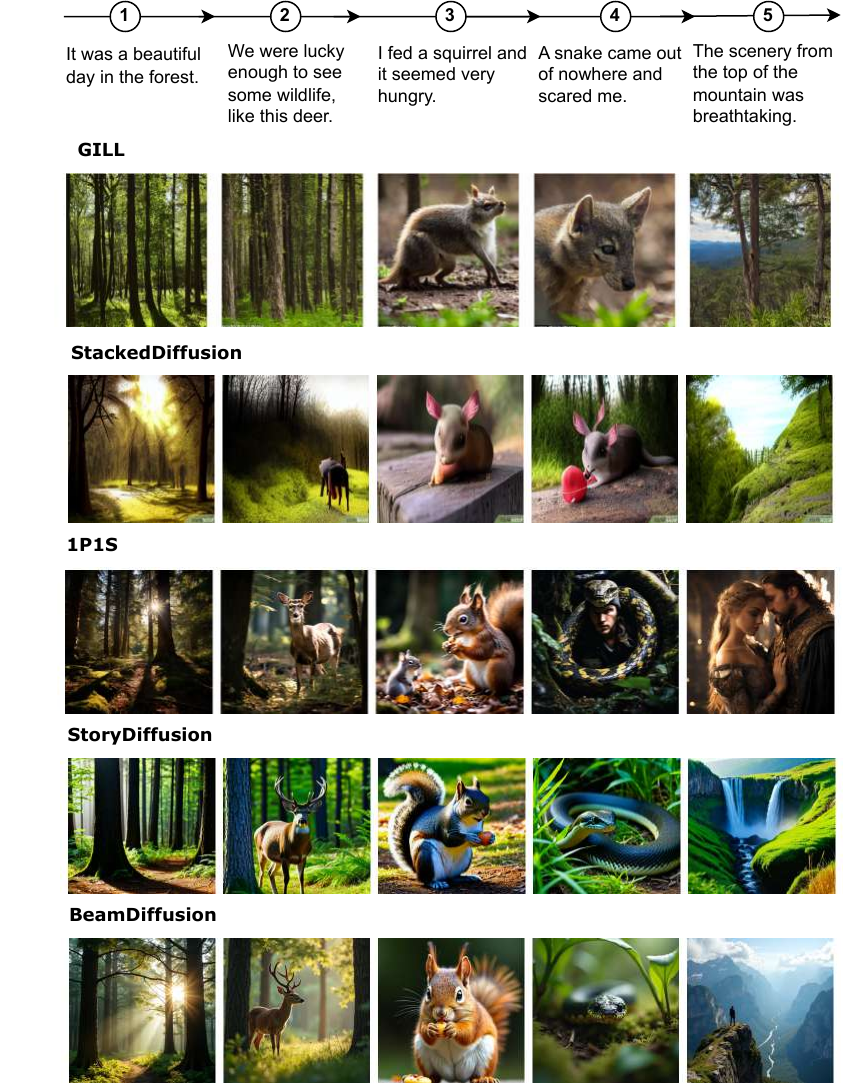} 
        \caption{Example of the VIST story domain \textit{"Yosemite Weekend"}, with a sequence of 5 steps.}
        \label{fig:vist_3}
    \end{minipage}
\end{figure*}

To provide a comprehensive comparison between all models, we present several examples across different domains in Figures~\ref{fig:recipe_13}-\ref{fig:vist_3}, highlighting how each method maintains sequence coherence in sequential image generation.

For the Recipes domain (\textit{"Buttermilk chicken wings"}, Figure~\ref{fig:recipe_13}), most models incorrectly depict the chicken as already cooked in step 2, instead of starting raw and gradually appearing cooked. \model correctly captures this progression, faithfully reflecting the cooking process across all steps. Similarly, in the Recipes task \textit{"Sweet potato and orange soup"} (Figure~\ref{fig:recipe_42}), GILL, StackedDiffusion, 1P1S, and StoryDiffusion misrepresent the ingredient being cut, while \model consistently depicts a person cutting a sweet potato and maintains stepwise visual consistency without hallucinations.

For the VIST stories, \model demonstrates strong identity and style preservation. In \textit{"4th of July 2006"} (Figure~\ref{fig:vist_4}), it is the only method to maintain the same child between step 2 and step 4. In \textit{"Yosemite Weekend"} (Figure~\ref{fig:vist_3}), both \model variants maintain coherent tree styles across steps, whereas other models show a noticeable break between step 1 and step 2.

Overall, while baseline methods may partially adhere to textual instructions or maintain visual similarity, \model consistently balances stepwise textual alignment and visual coherence, producing sequences that are both contextually accurate and visually stabl

\end{document}